\documentclass[10pt,twocolumn,letterpaper]{article}

\usepackage[pagenumbers]{cvpr} 

\usepackage[dvipsnames]{xcolor}

\definecolor{cvprblue}{rgb}{0.21,0.49,0.74}
\usepackage[pagebackref,breaklinks,colorlinks,citecolor=cvprblue]{hyperref}
\usepackage{adjustbox}
\usepackage{algorithm}
\usepackage{algpseudocode}
\usepackage{times}
\usepackage{epsfig}
\usepackage{graphicx}
\usepackage{amssymb}
\usepackage{booktabs}
\usepackage{amsmath,multirow,amsfonts,booktabs,xcolor}
\usepackage{array}
\usepackage{textcomp}
\usepackage{stfloats}
\usepackage{url}
\usepackage{verbatim}
\usepackage{colortbl}
\usepackage{xspace}
\usepackage{amsthm,enumitem}

\usepackage{rotating}

\title{UniHPE: Towards Unified Human Pose Estimation via Contrastive Learning}

\author{
Zhongyu Jiang$^{1}$ \quad 
Wenhao Chai$^1$ \quad 
Lei Li$^{2}$ \quad 
Zhuoran Zhou$^{1}$ \\
Cheng-Yen Yang$^1$ \quad 
Jenq-Neng Hwang$^1$ \\
[2mm]
University of Washington$^1$ \quad 
University of Copenhagen$^2$\\
[2mm]
\tt\small \{zyjiang, wchai, zhouz47, cycyang, hwang\}@uw.edu, lilei@di.ku.dk
}

\begin{document}
\maketitle
\begin{abstract} 
In recent times, there has been a growing interest in developing effective perception techniques for combining information from multiple modalities. This involves aligning features obtained from diverse sources to enable more efficient training with larger datasets and constraints, as well as leveraging the wealth of information contained in each modality. 2D and 3D Human Pose Estimation (HPE) are two critical perceptual tasks in computer vision, which have numerous downstream applications, such as Action Recognition, Human-Computer Interaction, Object tracking, \etc. Yet, there are limited instances where the correlation between Image and 2D/3D human pose has been clearly researched using a contrastive paradigm. In this paper, we propose \textbf{UniHPE}, a unified Human Pose Estimation pipeline, which aligns features from all three modalities, i.e., 2D human pose estimation,  lifting-based and image-based 3D human pose estimation, in the same pipeline. To align more than two modalities at the same time, we propose a novel singular value based contrastive learning loss, which better aligns different modalities and further boosts the performance. In our evaluation, UniHPE achieves remarkable performance metrics: MPJPE $50.5$mm on the Human3.6M dataset and PAMPJPE $51.6$mm on the 3DPW dataset.  Our proposed method holds immense potential to advance the field of computer vision and contribute to various applications.

\end{abstract}    
\section{Introduction}
\label{sec:intro}

Estimating 2D and 3D human poses (\ie, human keypoints) from only RGB images is one of the foundational tasks in the computer vision field, which can be further used for several downstream tasks like multiple object tracking~\cite{andriluka2018posetrack, snower2020keytrack}, action recognition~\cite{shahroudy2016nturgbd, duan2022pyskl}, human-computer interaction~\cite{wang2018hci, hu2018vgpn}, human body reconstruction~\cite{weng2022humannerf, li2023hybrik}, sports application~\cite{zhao2023survey, jiang2022golfpose}, \etc. Previous works follow the paradigms which estimate 3D human poses from 2D human poses (so-called lifting)~\cite{martinez2017simplebaseline, pavllo2019videopose3d, zhao2019semgcn, zheng2021poseformer, gong2021poseaug} or directly regress 3D human poses from RGB images (image-based)~\cite{kolotouros2019spin, georgakis2020hierarchical, sun2021ROMP, sun2022BEV, zhang2021pymaf, zhang2023pymafx}. Lifting networks learn the mapping between 2D and 3D human poses, and the image-based methods take advantage of the rich image information to get accurate pose estimation results. Yet, the lifting method remains a two-stage paradigm heavily reliant on the efficacy of 2D human pose detectors in the first phase. On the other hand, the absence of large-scale and diverse image-3D pose pairs data for training hurts the generalization of image-based methods. This brings up the question: \textit{Can we establish a unified human pose estimation paradigm by taking advantage of both lifting and image-based methods?}


\begin{figure}[t]
    \centering
    \includegraphics[width=0.8\linewidth]{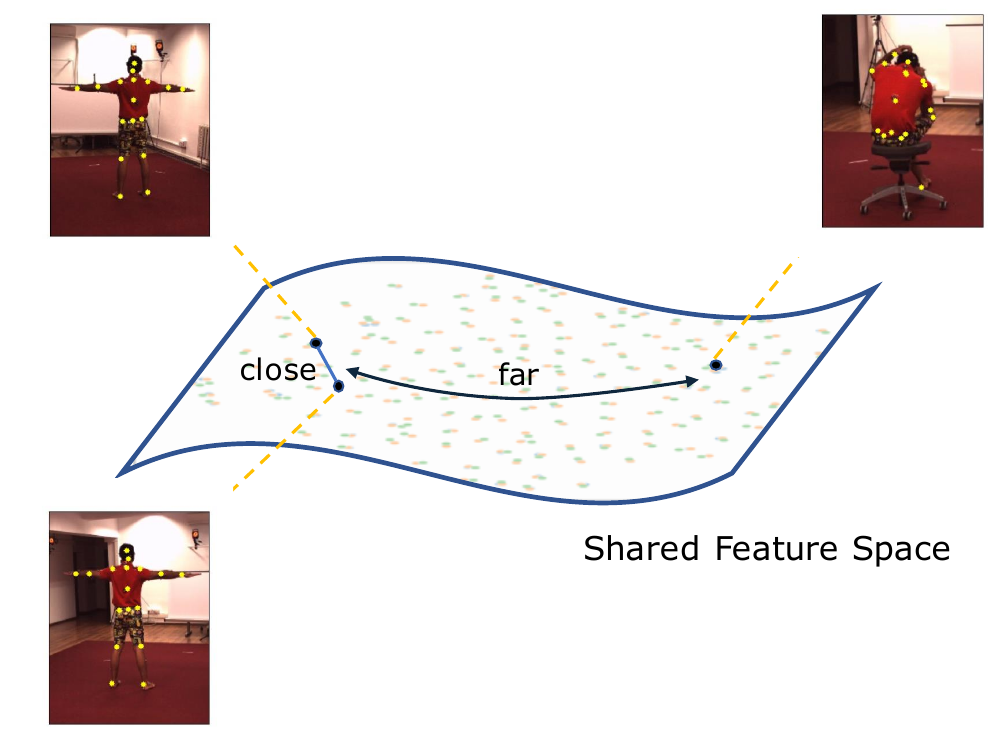}
    \caption{RGB image, 2D and 3D human pose embeddings extracted by corresponding encoders in the shared feature space. We show that after conducting contrastive learning during the pre-training stage, the embeddings of different modalities from the same sample are close to each other and away from other negative samples.}
    \label{fig:intro}
\end{figure}

From the perspective of representation learning, previous methods have been dedicated to mapping the representation of RGB images or 2D poses into the corresponding 3D pose space. In this paper, we propose UniHPE, a \underline{Uni}fied \underline{H}uman \underline{P}ose \underline{E}stimation framework, which aims to align RGB image, 2D and 3D human pose in the shared feature space in the pre-training stage, to further benefit pose estimation tasks. UniHPE can conduct 2D human pose estimation, lifting-based 3D human pose estimation, and image-based human pose estimation in a single model.

Learning joint embeddings across more than two modalities is a challenge. Inspired by Contrastive Language-Image Pre-Training (CLIP)~\cite{radford2021clip}, which proposes to learn transferable visual features with natural language supervisions trained on web-scale image-text pairs data, we claim that RGB image, 2D and 3D human pose representation alignment can also benefit from contrastive learning on large-scale and diverse datasets~(\eg, Human3.6M~\cite{ionescu2013h36m}, MPI-INF-3DHP~\cite{mono-3dhp2017}, \etc). 

Our proposed framework follows the encoder-decoder architecture, consisting of image, 2D and 3D human pose encoders, and 2D and 3D human pose decoders. The embedding features of these three modalities are shared in the bottleneck. To be specific, we first encode the images by HRNet~\cite{wang2020deep}, 2D and 3D human poses by Transformer~\cite{vaswani2017attention} respectively to get the corresponding embeddings. During the pre-training stage, we conduct contrastive learning to force the embeddings of different modalities from the same training sample to be close in the shared feature space. However, aligning more than two modalities is challenging, and therefore, we propose a singular value based contrastive learning loss to align three modalities at the same time. After that, during the training stage, we jointly train encoders and decoders with contrastive learning and multi-task learning simultaneously. During inference, since the embeddings are aligned in the same feature space, UniHPE supports 2D human pose estimation and lifting-based and image-based 3D human pose estimation in the same pipeline.

Our contributions can be summarised as follows:
\begin{itemize}
    \vspace{2pt}
    \item UniHPE can conduct both 2D and 3D HPE tasks in a single model and take advantage of both lifting and image-based 3D HPE methods.
    \vspace{2pt}
    \item We show that HPE tasks can benefit from feature alignment pre-training via contrastive learning.
    \vspace{2pt}
    \item We propose singular value based InfoNCE loss for contrastive learning to align more than two modalities at the same time.
    \vspace{2pt}
    \item UniHPE achieves MPJPE $50.5$ mm on the Human3.6M dataset and PAMPJPE $51.6$ mm on the 3DPW dataset.
\end{itemize}

\section{Related Works}
\label{sec:related}

Traditionally, the estimation of 3D human poses~\cite{wang2021deep} has been achieved by mainly two approaches: 2D-3D lifting and image-based method.

\subsection{Lifting Method for 3D HPE}

 2D-3D lifting~\cite{martinez2017simplebaseline, pavllo2019videopose3d, zhao2019semgcn, gong2021poseaug, chai2023global, jiang2023zedo, zhou2023zedoi} methods aim to infer 3D human pose under the assistance of 2D joint detector. Thus, the relations between 2D and 3D human poses have captivated the attention of numerous researchers in computer vision and human motion analysis. Though the internal correspondence is tight, it is rather challenging to connect their representations in the same embedding space as they contain various spatial information, and ambiguities in depth may also cause severe one-to-many 2D-3D mappings. 

\subsection{Image-based Method for 3D HPE}

The other approach for estimating 3D human poses is building an end-to-end network designed to predict the 3D joint coordinates of the poses or SMPL\cite{SMPL:2015} parameters directly from RGB images. Those methods can be categorized into two main classes: heatmap-based~\cite{pavlakos2017coarse, luvizon20182d} and regression-based~\cite{pavlakos2018learning, kolotouros2019spin, kocabas2020vibe, kocabas2021pare, zhang2021pymaf, li2022cliff, li2023hybrik} methods. Following the architecture of 2D human pose estimation, heatmap-based methods generate a 3D likelihood heatmap for each individual joint, and the joint's position is ascertained by identifying the peak within the heatmap. On the other hand, the regression-based methods detect the root location and regress the relative locations of other joints in two branches. In contrast, the SMPL regression methods focus on regressing SMPL parameters from image or video input. Kolotouros \etal~\cite{kolotouros2019spin} propose SPIN, which takes advantage of an optimization-based 3D pose estimation method, i.e., SMPLify~\cite{bogo2016smplify}, to achieve semi-supervised learning on 2D pose only datasets. VIBE~\cite{kocabas2020vibe} utilizes temporal information and a discriminator pretrained on a large 3D pose dataset to enhance the performance.


\subsection{Multi-Modal Learning and Feature Alignment}
How to align data from multiple modalities is a challenging and important task. CLIP~\cite{radford2021clip} is trained on web-scale text-image pairs under a contrastive learning paradigm. Inspired by the outstanding capability of learning representations for both vision and language of CLIP, various models have adopted the contrastive method to pursue zero-shot performance in other areas, such as~\cite{girdhar2023imagebind,guzhov2022audioclip,lin2022frozen,luo2022clip4clip}, by pre-training the model which maximizes cross-modal similarity scores. In this process, CLIP-based models would automatically learn implicit multi-modal alignments, which intensively reduces the difficulty of manually building feature correspondence. However, the majority of prior research has predominantly concentrated on visual-language or other visual-related cross-modal capacity, primarily due to the substantial availability of image-text paired datasets. Nonetheless, few works focus on the area of human pose. MPM~\cite{zhang2023mpm} aims to learn shared 2D-3D human pose features by the masked modeling paradigm. Yet, there are limited instances where the correlation between 2D/3D human pose has been clearly researched using a contrastive paradigm similar to our method. Furthermore, we further align image features with 2D and 3D human pose features and construct our unified human pose estimation pipeline.

\begin{figure*}[t]
    \centering
    \includegraphics[width=0.9\linewidth]{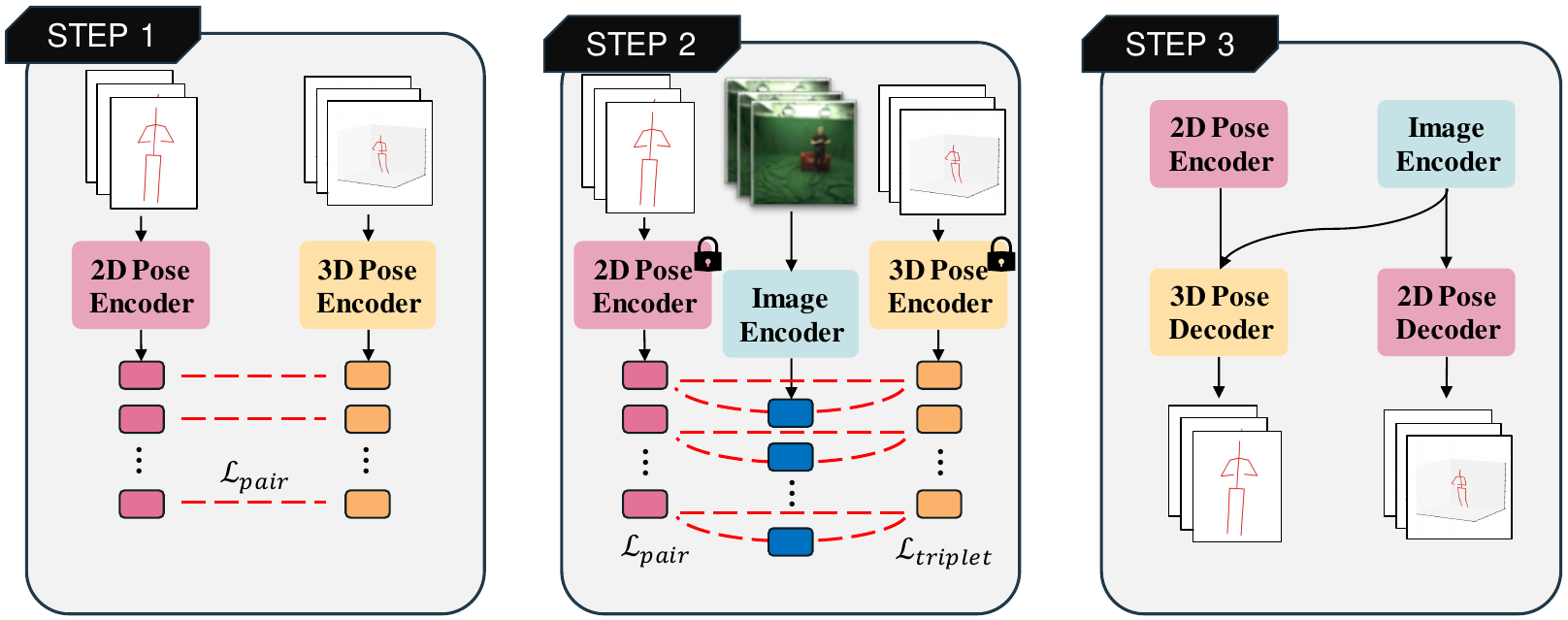}
    \caption{The training scheme of \textbf{UniHPE}. Steps 1 and 2 are pre-training stages, and Step 3 is for multi-task learning. During Step 1, we train the 2D and 3D pose embedding alignment first with $\mathcal{L}_{pair}$, and in Step 2, the image encoder is aligned with frozen 2D and 3D pose encoders via $\mathcal{L}_{pair}$ and $\mathcal{L}_{triplet}$. In Step 3, encoders and decoders are trained jointly via both contrastive learning and multi-task learning.}
    \label{fig:pipeline}
\end{figure*}

\begin{figure}[t]
    \centering
    \includegraphics[width=0.9\linewidth]{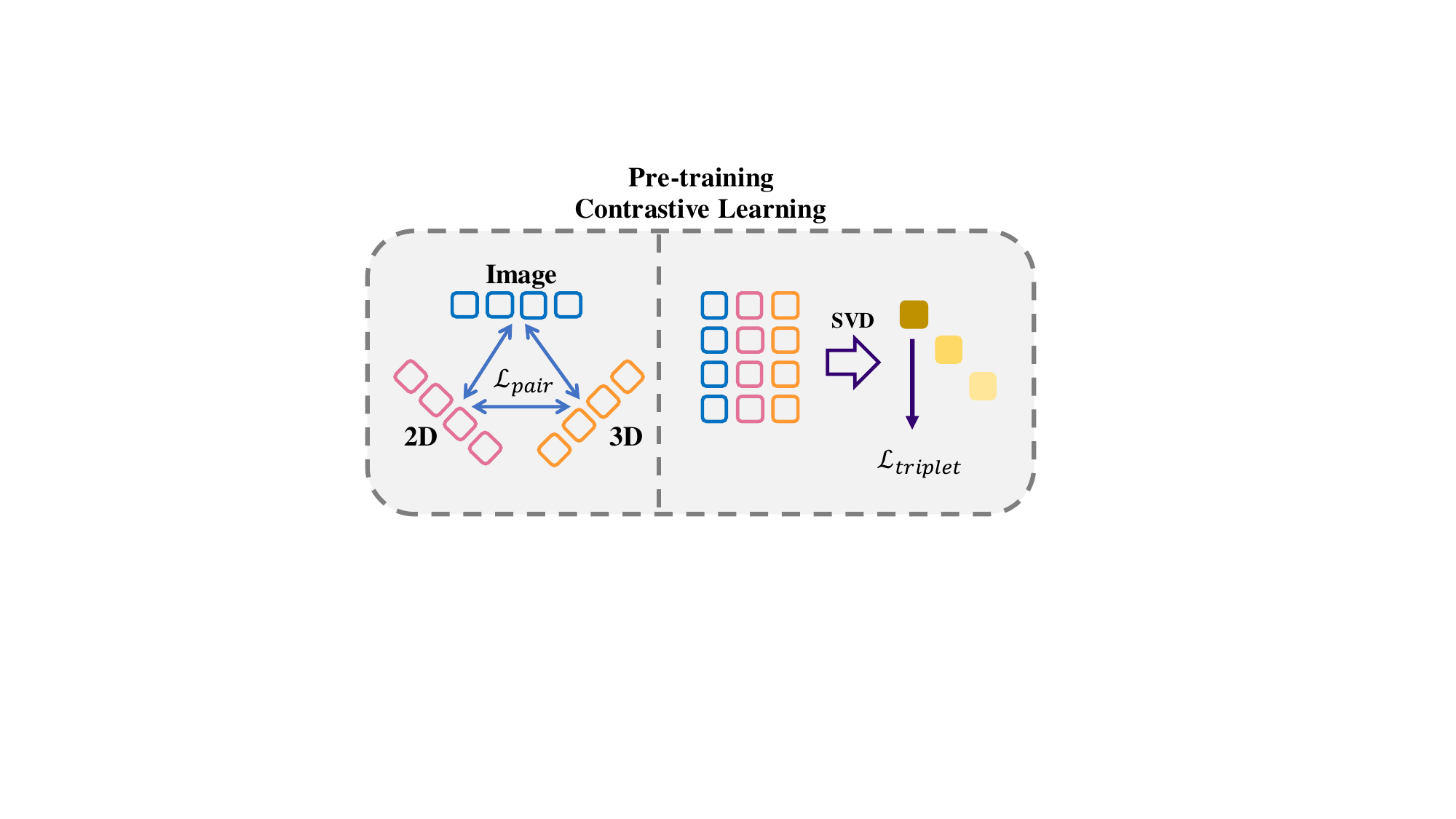}
    \caption{$\mathcal{L}_{pair}$ is applied three times for contrastive learning and the singular value based $\mathcal{L}_{triplet}$ focuses on aligning three modalities at the same time.}
    \label{fig:loss}
\end{figure}

\section{Methodology}
\label{sec:method}

We build a unified human pose estimation pipeline. During training, for any triplet of the cropped human image, $I \in \mathbb{N}^{H \times W \times 3}$, 2D and 3D human pose, $P_{2D/3D} \in \mathbb{R}^{J \times 2/3}$, UniHPE aligns the embeddings from all three modalities and utilizes 2D and 3D pose decoders for downstream tasks.

\subsection{Framework Architecture}

\paragraph{Image encoder.}
We simply use HRNet~\cite{wang2020deep} to extract the embeddings from RGB images, which is a convolution-based backbone for various visual recognition tasks. We concatenate and flatten the average pooled feature maps from the last stage and pass it through a linear projection layer to get a 1-D embedding vector as our image embedding.

\paragraph{2D/3D pose encoder.}
We adopt Transformer-based~\cite{vaswani2017attention} encoders to extract the embeddings from 2D and 3D human poses. We conduct bounding box normalized keypoint-wise patch embedding and retain the spatial information of each keypoint via adding learnable spatial position embedding. Then, the pose tokens prepended with a $\left[CLS\right]$ token and a bounding box token are fed into standard transformer encoder layers, including multi-head self-attention, feed-forward layers, and normalization layers. After that, we use the $\left[CLS\right]$ tokens as the 2D and 3D human pose embedding, respectively, which aggregates the information of the other tokens and can be regarded as general prior.

\paragraph{2D and 3D pose decoder.}
We build diffusion-based modules, $\mathcal{D}_{2D/3D}$, with two residual blocks to decode the embeddings and get 2D and 3D human poses. We treat the decoders following the Score Matching paradigm \cite{SMN} instead of DDPM\cite{ddpm} or DDIM\cite{ddim}. To be specific, the encoded embeddings are added with time embedding as well as a modality token, which indicates the source of the embedding (\eg from an image, 2D or 3D pose) in the diffusion network as condition embeddings and are used to generate the final 2D and 3D poses.  


We also try different architectures of encoders and decoders in the ablation study Section~\ref{sec:ablation}.

\subsection{Pre-training via Contrastive Learning}
During the pre-training stage, we aim to align the embeddings from images, 2D and 3D human poses via contrastive learning.  Given a batch of data, we have the RGB images, $I \in \mathbb{N}^{B \times H \times W \times 3}$, 2D poses, $P_{2D} \in \mathbb{R}^{B \times J \times 2}$, and 3D poses, $P_{3D} \in \mathbb{R}^{B \times  J \times 3}$, where $B, H, W, J$ are batch size, image height and width, and number of joints. We aim to train the image, 2D, and 3D pose encoders $E_{img}, E_{2D}, E_{3D}$ by maximizing the similarity between image embeddings $x_{img} \in \mathbb{R}^{B \times D}$, 2D pose embeddings $x_{2D} \in \mathbb{R}^{B \times D}$, and 3D pose embeddings $x_{3D} \in \mathbb{R}^{B \times D}$, where $D$ is the dimension of the embeddings, which is the same over three modalities. The most intuitive approach to aligning three modalities is to apply three pair-wise contrastive loss functions. 
For embeddings, $x_{\mathcal{S}}, x_{\mathcal{T}}$, from any pair of modalities, the contrastive learning loss is

\begin{figure*}[t]
\setcounter{figure}{0}
\renewcommand\figurename{Algorithm}
\begin{minipage}[b]{0.49\textwidth}
\begin{algorithm}[H]
    \centering
    \caption{Random Sampling in $\mathcal{L}_{svd}$}\label{algorithm}
    \begin{algorithmic}
        \Require \text{Image embeddings, $x_{img}$, 2D pose }
        \State \text{embeddings, $x_{2D}$, 3D pose embeddings, $x_{3D}$}
        \State \text{Batch size, $B$}
        \vspace{4pt}
        \State $\text{index}\_\text{list} \gets \text{zeros}(B, B, 3)$
        \For{$i\gets 0, B$}
        \State $\text{index}\_\text{list}\left[:,i,0 \right] \gets \text{arange}(B)$
        \State $\text{index}\_\text{list}\left[:,0,i \right] \gets \text{arange}(B)$
        \EndFor
        \For{$i\gets 1, B$}
            \For{$j\gets 1, 3$}
                \State $\text{index}\_\text{list}\left[:,i,j \right] \gets \text{shuffle}(\text{arange}(B))$
            \EndFor
        \EndFor
    \end{algorithmic}
\end{algorithm}
\end{minipage}
\hfill
\begin{minipage}[b]{0.49\textwidth}
\begin{algorithm}[H]
    \centering
    \caption{Implementation of $\mathcal{L}_{svd}$}\label{algorithm1}
    \begin{algorithmic}
        \Require \text{Normalized image embeddings, $x_{img}$ 2D pose}
        \State \text{embeddings, $x_{2D}$, 3D pose embeddings, $x_{3D}$,}
        \State \text{Batch size, $B$, Temperature, $\tau$} \vspace{1pt}
        \State $\mathcal{M}_x \gets \text{stack}((x_{img}, x_{2d}, x_{3d}), dim=1)$
        \State $\text{index}\_\text{list} \gets \text{RandSample}(x_{img}, x_{2D}, x_{3D})$
        \State {\color{gray}\text{\# $\mathcal{M}_x \in \mathbb{R}^{B \times B \times 3 \times D}$}}
        \State $\mathcal{M}_x \gets \mathcal{M}_x\left[ \text{index}\_\text{list} \right]$
        \State {\color{gray}\text{\# $\mathcal{M}_x \in \mathbb{R}^{B \times B \times 3 \times 3}$}}
        \State $\mathcal{M}_x \gets \mathcal{M}_x\mathcal{M}_x^\intercal$
        \State $\text{logits} \gets \text{eigenval}(\mathcal{M}_x)\left[:,:,0\right]$
        \State $\text{logits} \gets \text{logits} / \tau$
        \State $\text{label} \gets \text{zeros}(B)$
        \State $\mathcal{L}_{triplet} = \text{CrossEntropy}(\text{logits}, \text{label})$
    \end{algorithmic}
\end{algorithm}
\end{minipage}
\caption{The \textbf{pseudo-code} of triplet random sampling and the implementation of $\mathcal{L}_{triplet}$. To simplify the computation, for each mini batch, we randomly sample $B-1$ negative triplets and one positive triplet. In }
\label{alg:svdloss}
\end{figure*}

\begin{equation}
    \mathcal{L}_{pair} = -\log \frac{\exp{(x_{\mathcal{S}} \cdot x_{\mathcal{T}}^{+} / \tau)}}{\sum_{i=1}^{B}\exp(x_{\mathcal{S}} \cdot x_{\mathcal{T},i} / \tau)},
\end{equation}

where $\tau$ is the learnable temperature initialized by $\tau_0$.

However, we found that simply applying three pairwise InfoNCE loss cannot obtain expected embedding similarity across three modalities, as shown in the ablation studies Section~\ref{sec:ablation:loss}. Therefore, we propose a singular value-based InfoNCE loss (Triplet-InfoNCE) to address this issue. 

We stack the embeddings from three modalities to build a normalized embedding matrix formulated by

\begin{equation}
    \mathcal{M}_x = 
    \begin{bmatrix}
        x_{img} & x_{2D} & x_{3D}
    \end{bmatrix}^T \in \mathbb{R}^{3 \times D}.
\end{equation}

If we apply singular value decomposition (SVD) to this matrix, $M_x = U\Sigma V^*$, the largest singular value, $\sigma_1 = \Sigma_{11}$, is related to the linear correlation of row vectors. Meanwhile, since the embeddings are normalized, the largest singular value should be in $\left[-\sqrt{3}, \sqrt{3}\right]$. Therefore, we can use InfoNCE loss to align any triplet of embeddings by maximizing the $\sigma_1$. However, computing the singular value of a matrix with $3 \times D$, where $3 \ll D$, is time-consuming. In our implementation, the largest eigenvalue $\lambda_1$ of the matrix $\mathcal{M}_x\mathcal{M}_x^\intercal \in \mathbb{R}^{3 \times 3}$ is our optimization target, since $\lambda_1 = \sigma_1^2$. Therefore, by maximizing the $\lambda_1$ for positive triplets, which contain three embeddings from the same frame, and minimizing the $\lambda_1$ for negative triplets, which contain at least one embedding from a different frame, we are able to align embeddings from three modalities jointly.

However, in one mini batch, the number of negative triplets for any positive triplet is $3B^2 - 3B + 1$, and if we use all the negative samples as our denominator in InfoNCE loss, the time consumption is unacceptable. As shown in the Alg~\ref{alg:svdloss}, we apply a random sample algorithm to select only $B - 1$ negative triplets for each positive triplet. In this case, the singular value based InfoNCE loss can be formed as,

\begin{equation}
    \mathcal{L}_{triplet} = -\log \frac{\exp{(\lambda_1^{+} / \tau)}}{\sum_{i=1}^{B}\exp(\lambda_{1i} / \tau)}.
\end{equation}

Overall, our contrastive learning loss is

\begin{equation}
    \mathcal{L}_{cl} = \mathcal{L}_{pair} + \alpha \mathcal{L}_{triplet}.
\end{equation}
where $\alpha $ is the weighted factor.

\subsection{Training via Multi-Task Learning}
After the pre-training stage, all encoders and decoders are trained jointly. While encoders are trained with $\mathcal{L}_{cl}$, the task losses, $\mathcal{L}_{2D/3D}$, depend on the architectures of decoders. For diffusion-based decoders, we adopt the loss from the Score Matching Network~\cite{song2020score}, and for MLP-based decoders, we utilize simple mean square error losses.

Therefore, the overall loss is

\begin{equation}
    \mathcal{L} = \mathcal{L}_{cl} + \mathcal{L}_{2D} + \mathcal{L}_{3D}.
\end{equation}

During inference, since the embeddings are aligned in the same feature space, UniHPE can support 2D human pose estimation and lifting-based and image-based 3D human pose estimation in the same pipeline. The 2D human pose can be estimated by sending the image embedding to the 2D decoder and the 3D human pose can be predicted by sending 2D or image embedding to the 3D decoder.


\section{Experiments}
\subsection{Datasets and Performance Metrics}

\begin{table*}[t]
  \centering
  \resizebox{0.6\linewidth}{!}{
    \begin{tabular}{cl|c|cc}
    \toprule
    \multicolumn{2}{c|}{\multirow{2}[4]{*}{Method}} & {3DPW} & \multicolumn{2}{c}{Human3.6M} \\
\cmidrule{3-5}    \multicolumn{2}{c|}{} & PA-MPJPE~($\downarrow$)  & MPJPE~($\downarrow$)  & PA-MPJPE~($\downarrow$)  \\
    \midrule
    \multirow{4}[4]{*}{\begin{sideways}\rotatebox[origin=c]{0}{Temporal}\end{sideways}} 
    & VideoPose3D~(f=243)~\cite{pavllo2019videopose3d} & 68.0 & 46.8 & 36.5 \\
    & AdaptPose~\cite{gholami2022adaptpose} & 46.5 & - & - \\
    & Li \etal~\cite{li2022exploiting} & - & 43.7 & 35.2 \\
    & MixSTE~\cite{zhang2022mixste} & - & 40.9 & 32.6 \\
    & MPM~\cite{zhang2023mpm} & - & 42.6 & 34.7 \\
    \midrule
    \multirow{4}[4]{*}{\begin{sideways}\rotatebox[origin=c]{0}{Frame-based}\end{sideways}} 
    & SimpleBaseline~\cite{martinez2017simplebaseline} & 89.4 & 62.9 & 47.7 \\
    & SemGCN~\cite{zhao2019semgcn} & 102.0 & 61.2 & 47.7 \\
    & VideoPose3D~(f=1)~\cite{pavllo2019videopose3d} & 94.6 & 55.2 & 42.3 \\
    & PoseAug~\cite{gong2021poseaug} & 58.5 & \textbf{52.9} & - \\
    & PoseDA~\cite{chai2023global} & \underline{55.3} & - & - \\
    \rowcolor[gray]{0.9}
    & UniHPE~(ours) & \textbf{51.6} & \underline{53.6} & \textbf{40.9}\\
    \bottomrule
    \end{tabular}%
    }
    \caption{\textbf{Lifting-based 3D HPE} performance on the 3DPW and Human3.6M datasets under MPJPE and PA-MPJPE. The ground truth 2D keypoints are used on 3DPW dataset, while the detected 2D keypoints from CPN are used on Human3.6M dataset.
    }
  \label{tab:2d-3d}%
\end{table*}%
We use several widely used 3D human pose datasets to train and evaluate our framework, including Human3.6M~\cite{ionescu2013h36m}, MPI-INF-3DHP~\cite{mono-3dhp2017}, and 3DPW~\cite{von20183dpw}. We train UniHPE on Human3.6M and 3DHP datasets and evaluate it on Human3.6M and 3DPW datasets.

\paragraph{Human3.6M} Human3.6M~\cite{ionescu2013h36m} dataset, which contains 3.6 million frames of corresponding 2D and 3D human poses, including 5 female and 6 male subjects under 17 different scenarios, is a video dataset captured using a MoCap system. Following the previous works, we choose 5 subjects (S1, S5, S6, S7, S8) for training, and the other 2 subjects (S9 and S11) for evaluation. We report the Mean
Per Joint Position Error (MPJPE) as the performance metric of Protocol \#1 as well as Procrusts analysis MPJPE (PA-MPJPE) as the
metric of Protocol \#2 for both the lifting path and image-bath path in our proposed framework.

\paragraph{MPI-INF-3DHP}
Compared to Human3.6M, MPI-INF-3DHP~\cite{mono-3dhp2017} is a more challenging 3D human pose dataset captured indoors and outdoors. MPI-INF-3DHP contains 1.3 million frames, consisting of 4 males and 4 females with 8 types of action captured by 14 cameras covering a greater diversity of poses. We use MPI-INF-3DHP in training to enrich the training samples and enhance the performance of our image branch.

\paragraph{3DPW}
3DPW~\cite{von20183dpw} is the first dataset that includes video footage taken from a moving phone camera. It includes 60 video sequences as the training dataset with 22k images, while the testing dataset includes 35k images. Compared to Human 3.6M or MPI-INF-3DHP, 3DPW is a more challenging in-the-wild dataset, with uncontrolled motion and scene. Similar to most other works, we only evaluate our model under PA-MPJPE metrics, without considering MPJPE, due to the fact that the scale of the human body, camera intrinsic, and distance of 3DPW are not compatible with the training data.

\subsection{Implementation Details}

\paragraph{Training settings.} We implement our proposed framework using PyTorch~\cite{paszke2019pytorch} on a single NVIDIA A100 80G GPU. The pre-training includes two stages: (1) 2D-3D alignment; (2) Image-2D-3D joint alignment. In the first stage of pre-training, the batch size is $2048$, $\tau_0 = 1/14$, and $\tau \in \left[1/100, 10^4\right]$, while in the second stage, the batch size is $180$, $\tau_0 = 1/5$, and $\tau \in \left[1/10, 10^4\right]$. During the multi-task training stage, encoders and decoders are trained together with the batch size being $180$, $\tau_0 = 1/5$, and $\tau \in \left[1/5, 10^4\right]$. For the weight of triplet contrastive loss, $\mathcal{L}_{triplet}$, $\alpha = 1$. The input image size of the image encoder is $192\times256$. During both of the two stages, we adopt Adam optimizer with a learning rate of $1\times 10^{-4}$. We train UniHPE on Human3.6M~\cite{ionescu2013h36m} and MPI-INF-3DHP~\cite{mono-3dhp2017} datasets and apply ablation study about the performance difference of using different training datasets.

\subsection{Comparison with State-of-the-Art Methods}

\paragraph{Lifting-based 3D Human Pose Estimation}
We evaluate the performance of lifting-based 3D HPE tasks on Human3.6M and 3DPW datasets. As shown in Table~\ref{tab:2d-3d}, UniHPE archives $51.6$ mm in terms of PA-MPJPE on 3DPW dataset and $53.6$ mm in terms of MPJPE on Human3.6M dataset, which is the state-of-the-art performance. Since UniHPE is not trained on 3DPW, it is a fair comparison with those cross-domain evaluation methods.

\paragraph{Image-based 3D Human Pose Estimation}

\begin{table}[t]
  \centering
  \resizebox{\linewidth}{!}{
    \begin{tabular}{cl|c|cc}
    \toprule
    \multicolumn{2}{c|}{\multirow{2}[4]{*}{Method}} & 3DPW & \multicolumn{2}{c}{Human3.6M} \\
\cmidrule{3-5}    \multicolumn{2}{c|}{} & PA-MPJPE~($\downarrow$)  & MPJPE~($\downarrow$)  & PA-MPJPE~($\downarrow$)  \\
    \midrule
    \multirow{5}[2]{*}{\begin{sideways}\rotatebox[origin=c]{0}{Temporal}\end{sideways}} & Kanazawa~\etal~\cite{kanazawa2019learning} & 72.6 & -     & 56.9 \\
          & Doersch~\etal~\cite{doersch2019sim2real} & 74.7  & -     & - \\
          & Arnab~\etal~\cite{arnab2019exploiting} & 72.2 & 77.8  & 54.3 \\
          & DSD~\cite{sun2019dsd} & 69.5  & 59.1  & 42.4 \\
          & VIBE~\cite{kocabas2020vibe} & 56.5 & 65.9  & 41.5 \\
    \midrule
    \multirow{16}[4]{*}{\begin{sideways}\rotatebox[origin=c]{0}{Frame-based}\end{sideways}} & Pavlakos~\etal~\cite{pavlakos2018learning} & -     & -     & 75.9 \\
          & HMR~\cite{kanazawa2018hmr} & 76.7 & 88.0    & 56.8 \\
          & NBF~\cite{omran2018nbf} & -     & -     & 59.9 \\
          & GraphCMR~\cite{kolotouros2019cmr} & 70.2 & -     & 50.1 \\
          & HoloPose~\cite{guler2019holopose} & -     & 60.3  & 46.5 \\
          & DenseRaC~\cite{xu2019denserac} & -     & 76.8  & 48.0 \\
          & SPIN~\cite{kolotouros2019spin} & 59.2 & 62.5  & 41.1 \\
          & DecoMR~\cite{zeng2020decomr} & 61.7  & -     & 39.3 \\
          & HKMR~\cite{georgakis2020hkmr} & -     & 59.6  & 43.2 \\
          & PyMAF~\cite{zhang2021pymaf} & 58.9         & 57.7         & 40.5         \\
          & PARE~\cite{kocabas2021pare} & 50.9 & 76.8         & 50.6         \\
          & PyMAF-X~\cite{zhang2023pymafx} & \underline{47.1} & 54.2 & 37.2 \\
          & CLIFF~\cite{li2022cliff} & \textbf{43.0} & \textbf{47.1} & \textbf{32.7} \\
          \rowcolor[gray]{0.9}
          & UniHPE-w32~(ours) & 67.1 & 55.2 & 40.0\\
          \rowcolor[gray]{0.9}
          & UniHPE-w48~(ours) & 65.7 & \underline{50.5} & \underline{36.2} \\
    \bottomrule
    \end{tabular}%
    }
    \caption{\textbf{Image-based 3D HPE} performance on the 3DPW and Human3.6M datasets under MPJPE and PA-MPJPE.
    }
  \label{tab:image-3d}%
\end{table}%
As for image-based 3D HPE, we also evaluate the performance on Human3.6M and 3DPW datasets. As shown in Table~\ref{tab:image-3d}, UniHPE archives $50.5$ mm and $36.2$ mm in terms of MPJPE and PA-MPJPE on Human3.6M dataset, as well as $65.7$ mm of PA-MPJPE on 3DPW dataset. Note that we are the only keypoint-based method in the Table~\ref{tab:image-3d}, and all the others are SMPL-based. UniHPE archives competitive results regarding the number of model parameters and training data with SOTA methods, showing the effectiveness of our method.

\begin{figure*}[t]
    \setcounter{figure}{3}
    \centering
    \includegraphics[width=0.9\linewidth]{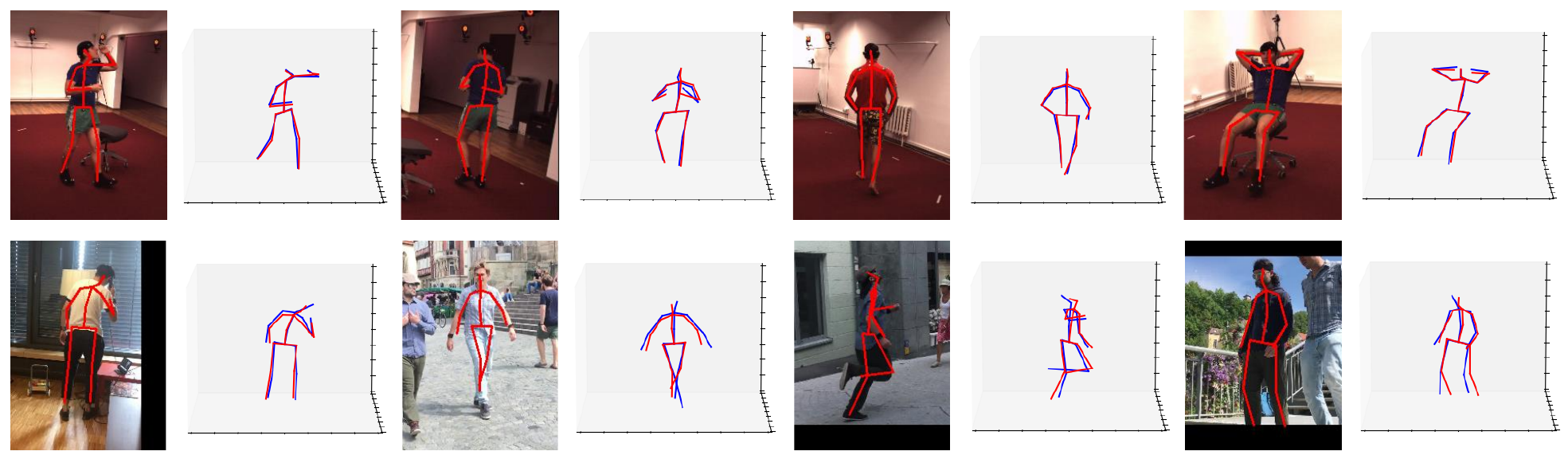}
    \caption{Qualitative results on Human3.6M (row 1) and 3DPW (row 2) datasets. Red poses are ground truth, and the blue ones are estimated from the image branch.}
    \label{fig:vis}
\end{figure*}

\paragraph{2D Human Pose Estimation}
\begin{table}[t]
\centering
\resizebox{0.6\linewidth}{!}{
\begin{tabular}{l|cc}
\toprule
Model & PCK$_{0.05}$~($\uparrow$) & EPE~($\downarrow$) \\
\midrule
HRNet-w32 & 91.1 & 9.43  \\
HRNet-w48 & 93.2 & 7.36 \\
\midrule
\rowcolor[gray]{0.9}
UniHPE-w32 & \textbf{93.5} & \textbf{7.18} \\
\bottomrule
\end{tabular}
}
\caption{2D Pose Estimation on Human3.6M dataset. UniHPE with HRNet-w32 as the image encoder outperforms the original HRNet-w32 and even HRNet-w48.}
\label{tab:image-2d}
\end{table}
Since our pipeline also supports 2D pose estimation, we compare the 2D pose estimation performance of UniHPE with the SOTA method on the Human3.6M dataset. The evaluation metrics are Percentage of Correct Keypoints (PCK) with a threshold as $0.05$ and End Point Errors (EPE). As shown in Table~\ref{tab:image-2d}, the UniHPE with HRNet-w32 as backbone can even outperform the original HRNet-w48 with PCK$_{0.05}$ as $93.5$ and EPE as $7.18$.

\subsection{Ablation Study}
\label{sec:ablation}
In this section, we conduct extensive ablation studies to investigate the importance of each module in the UniHPE, especially how the singular value based loss, $\mathcal{L}_{triple}$, helps the training. 

\begin{figure*}[t]
  \centering
  \subfloat[2D-3D pose]
  {
      \includegraphics[width=0.32\textwidth]{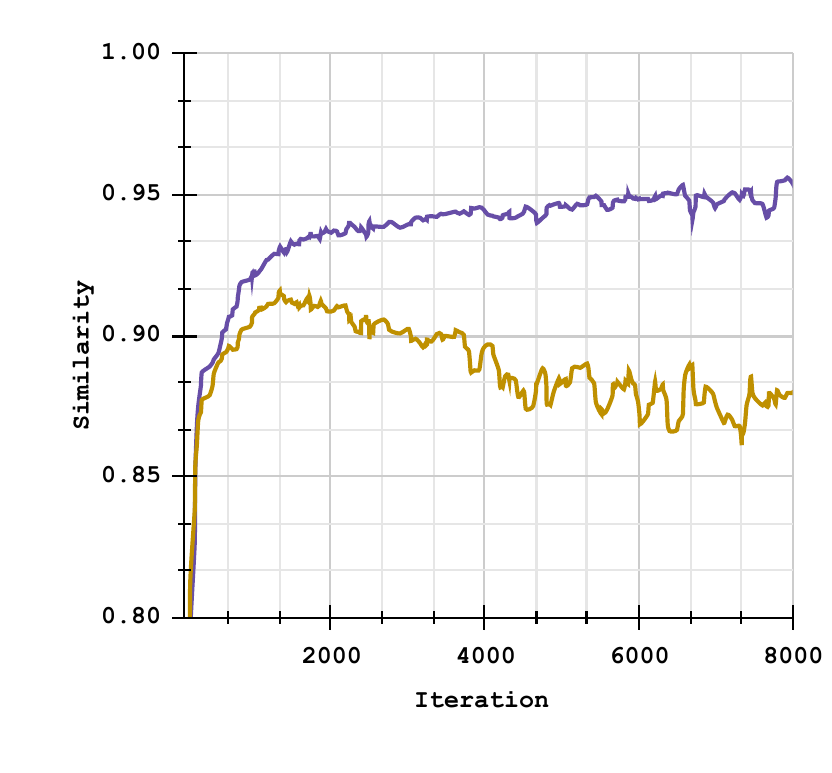}
  }
  \subfloat[image-2D pose]
  {
      \includegraphics[width=0.32\textwidth]{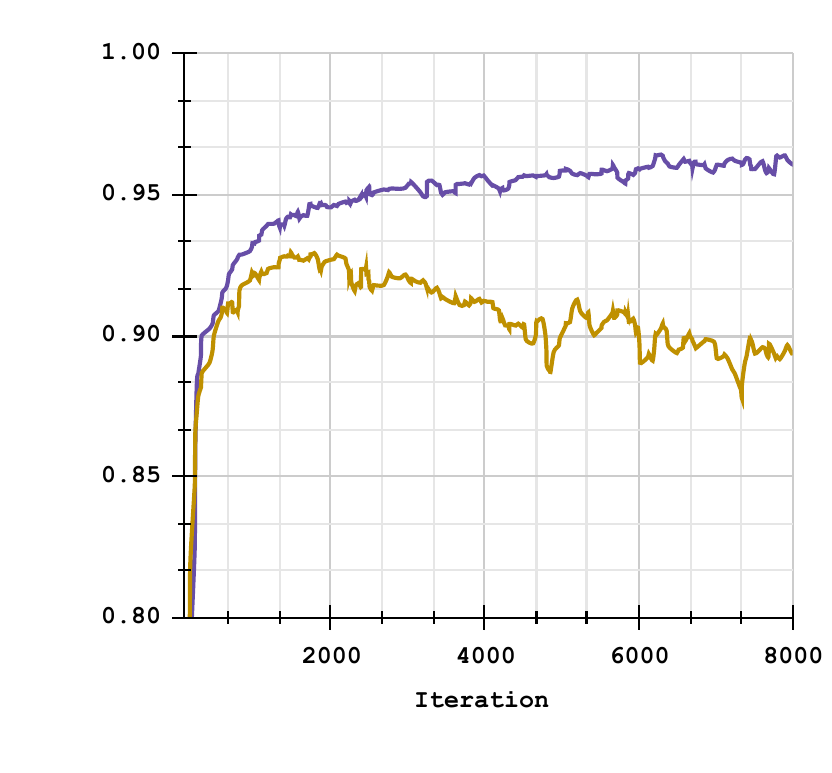}
  }
  \subfloat[image-3D pose]
  {
      \includegraphics[width=0.32\textwidth]{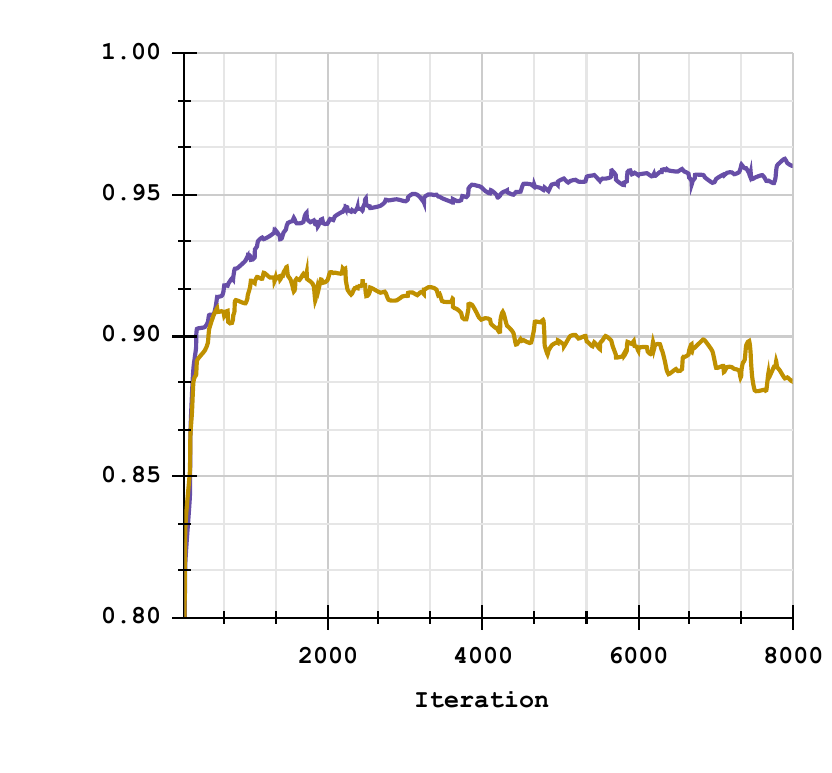}
  }
  \caption{\textbf{Cosine similarities} between different modalities. The yellow line is the one only trained with three pair-wise losses, $\mathcal{L}_{pair}$, and the purple line is the training curve with additional singular value-based InfoNCE loss, $\mathcal{L}_{triplet}$. Our proposed singular value-based InfoNCE loss helps align the feature space.}
  \label{fig:ablation_loss} 
\end{figure*}

\paragraph{Encoder and decoder architecture.}
\begin{table}[t]
\centering
\resizebox{0.9\linewidth}{!}{
\begin{tabular}{ll|cc}
\toprule
2D Encoder  & 3D Decoder & MPJPE~($\downarrow$)  & PA-MPJPE~($\downarrow$) \\ 
\midrule
MLP  &   MLP  &   60.5  & 44.5 \\
transformer & MLP & 64.8 & 48.1   \\
MLP  &   diffusion  &    71.4 & 45.6   \\
\rowcolor[gray]{0.9}
transformer  &   diffusion  &   \textbf{39.3} & \textbf{29.9}  \\
\bottomrule
\end{tabular}
}
\caption{\textbf{Ablation study} on architecture selection in lifting-based 3D HPE task. Evaluated on Human3.6M dataset with ground truth 2D keypoints.}
\label{tab:ablation_arch_lift}
\end{table}

\begin{table}[t]
\centering
\resizebox{0.75\linewidth}{!}{
\begin{tabular}{l|cc}
\toprule
image Encoder       & MPJPE~($\downarrow$)  & PA-MPJPE~($\downarrow$) \\ 
\midrule
ViT-S & 151.6	& 93.8 \\
ViT-B & 149.9 &  93.4 \\
HRNet-w32 & 55.2  & 40.0 \\	
\rowcolor[gray]{0.9}
HRNet-w48 & \textbf{50.5} & \textbf{36.2} \\	
\bottomrule
\end{tabular}
}
\caption{\textbf{Ablation study} on image encoder architecture selection in image-based 3D HPE task. Evaluated on Human3.6M dataset.}
\label{tab:ablation_arch_image}
\end{table}
Since UniHPE is flexible with any encoder or decoder architecture, we conduct experiments on several strong backbones with different architectures, including MLP, CNN, transformer, and diffusion. We first select HRNet-w32 as the image encoder during alignment training and evaluate the performance of the lifting branch under different combinations of 2D pose encoder and 3D pose decoder architecture in the 3D HPE task on the Human3.6M dataset. As shown in Table~\ref{tab:ablation_arch_lift}, transformer-based 2D pose encoder and diffusion-based 3D pose encoder achieve the best performance. Then, we keep this selection and further conduct ablation studies in the image encoder part. As shown in Table~\ref{tab:ablation_arch_image},  HRNet-w48 outperforms then other encoders in the image-based 3D HPE task on the Human3.6M dataset. We utilize the coco pre-trained HRNet and ViTPose\cite{xu2022vitpose} as the initial weights for a fair comparison.

\paragraph{End-to-End training without alignment.}
\begin{table*}[t]
\newcommand{\R}[1]{~\scriptsize{\textcolor{red}{(#1)}}}
\newcommand{\G}[1]{~\scriptsize{\textcolor{OliveGreen}{(#1)}}}
\centering
\resizebox{0.85\linewidth}{!}{
\begin{tabular}{cccc|cc|cc}
\toprule
  & & & & \multicolumn{2}{c|}{GT 2D} & \multicolumn{2}{c}{Image} \\
 $\mathcal{L}_{pair}$ & $\mathcal{L}_{triplet}$ & $\mathcal{M}$ Token & +3DHP & MPJPE~($\downarrow$)  & PA-MPJPE~($\downarrow$) & MPJPE~($\downarrow$)  & PA-MPJPE~($\downarrow$) \\ 
\midrule
 & & & & 41.3 & 31.6 & 91.8 & 68.7\\
\midrule
\checkmark & & & & 60.0\R{+18.7} & 47.5\R{+15.9} & 65.5\G{-26.3} & 51.8\G{-16.9}\\
\checkmark & \checkmark & & & 40.9\G{-0.4} & 31.7\R{+0.1}  & 58.7\G{-33.1} & 44.4\G{-24.3}\\
\checkmark & \checkmark & \checkmark & & \textbf{39.3\G{-2.0}} & \textbf{29.9\G{-1.7}} & 57.5\G{-34.3} & 42.9\G{-25.8}\\	
\checkmark & \checkmark & \checkmark & \checkmark & 44.2\R{+2.9} & 33.6\R{+2.0} & \textbf{55.1\G{-36.7}} & \textbf{40.0\G{-28.7}} \\
\bottomrule
\end{tabular}
}
\caption{\textbf{Ablation study} on UniHPE. Evaluated on Human3.6M dataset. $\mathcal{L}_{pair}$ and $\mathcal{L}_{triplet}$ denotes applying those losses on the pre-training stage. $\mathcal{M}$ token means decoders utilize the modality token. We evaluate the performance with additional data from MPI-INF-3DHP dataset as well.}
\label{tab:ablation}
\end{table*}
We claim that feature alignment, pre-training via contrastive learning, between different modalities is the key to success. Therefore, we conduct the ablation studies on skipping the alignment training stages. As shown in Table~\ref{tab:ablation}, alignment improves the image-based 3D HPE performance significantly on the Human3.6M dataset. As shown in table~\ref{tab:ablation}, without contrastive learning, the performance gap between lifting and image branches shows that the features are not correctly aligned. Furthermore, the combination of $\mathcal{L}_{triplet}$ and $\mathcal{L}_{pair}$ provides the best performance on both lifting and image branches, which matches with the comparative results shown in Fig~\ref{fig:ablation_loss}. 

\paragraph{Ablation on modality token.}

In UniHPE, we design a modality token when using the 3D pose decoder to estimate 3D human poses. The modality token indicates which modality the features came from (\eg image or 2D pose). As shown in Table~\ref{tab:ablation},  consistent improvement is observed in using the modality token among lifting-based and image-based 3D HPE tasks on the Human3.6M dataset.

\paragraph{Training objective in pre-training.}
\label{sec:ablation:loss}
We conduct training objectives in the alignment training stage. As shown in Figure~\ref{fig:ablation_loss}, compared to simply applying three pairwise InfoNCE loss, $\mathcal{L}_{pair}$, the proposed singular value-based InfoNCE loss, $\mathcal{L}_{triplet}$, significantly better aligns the feature from different modality. With the help of $\mathcal{L}_{triplet}$, the embedding cosine similarity between different modalities does not decrease after around $1500$ iterations and keeps increasing to around $0.95$ in $8000$ iterations.

\paragraph{Training with additional data.}
As shown in Table~\ref{tab:ablation}, by training the UniHPE on an additional dataset, like 3DHP, the performance of the lifting-based 3D pose estimation drops, but the image-based branch improves. It is noted that the distribution of 2D and 3D pose pairs on 3DHP differs from Human3.6M, which increases the robustness of the lifting-based branch but decreases the performance on Human3.6M. However,  the training with additional data boosts the image-based branch by improving the image data diversity.

\begin{figure}[t]
    \centering
    \includegraphics[width=0.9\linewidth]{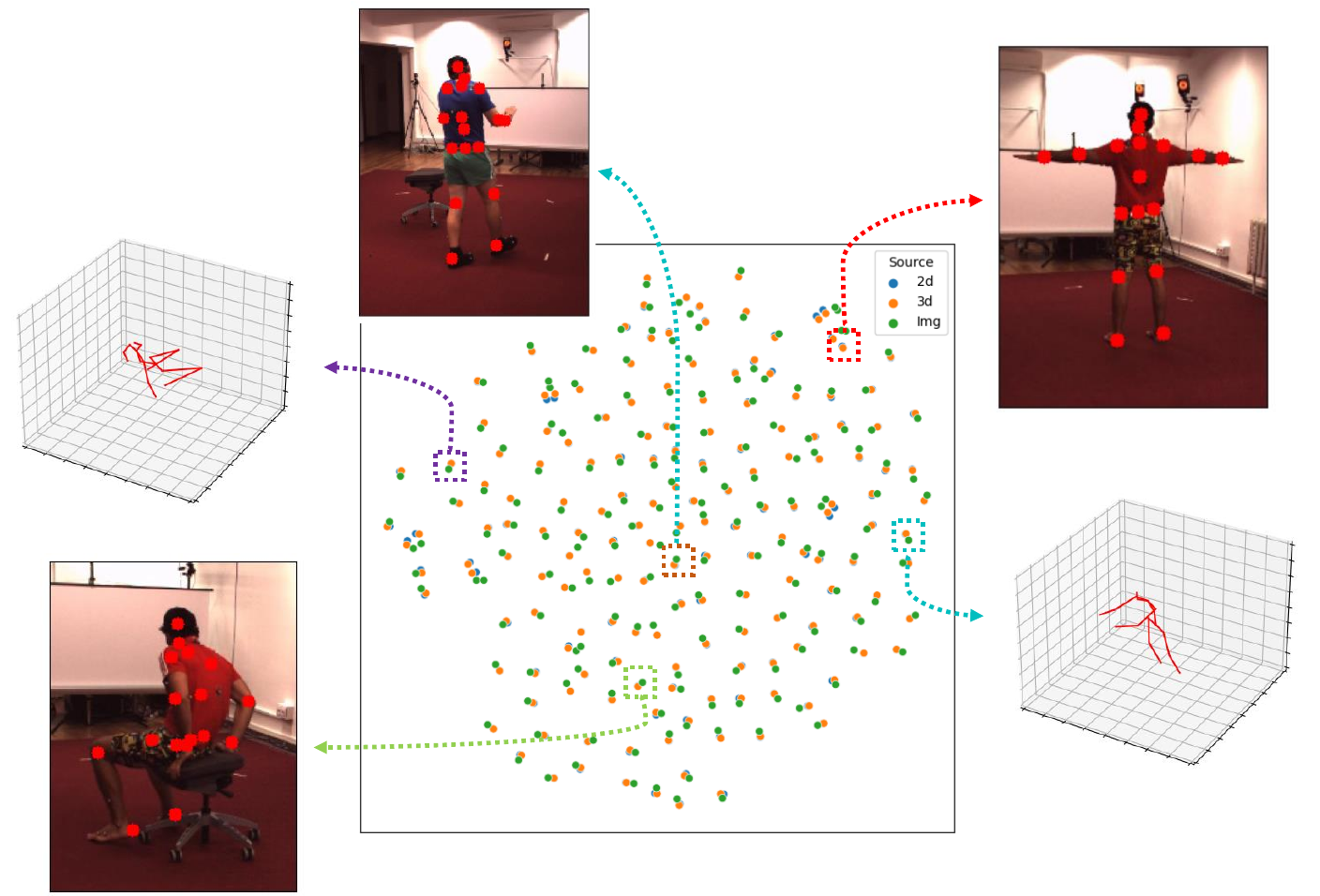}
    \caption{The visualization of the shared feature space from UniHPE using t-SNE. The features of different modalities are aligned with each other and uniformly distributed in the shared feature space.}
    \label{fig:feature}
\end{figure}
\begin{figure}[t]
    \centering
    \includegraphics[width=\linewidth]{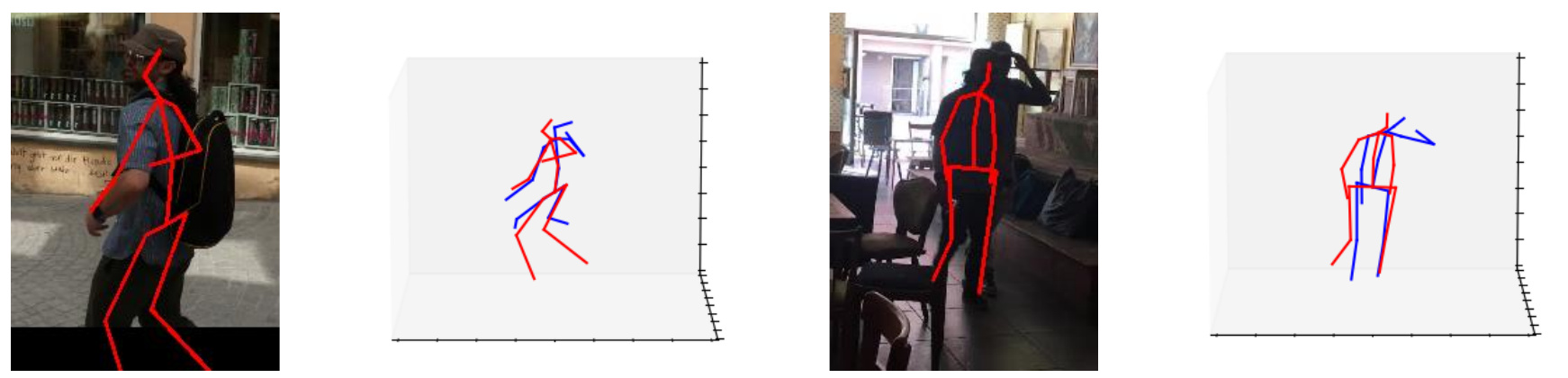}
    \caption{Failure cases of UniHPE. When there is heavy occlusion, our model may estimate the incorrect pose or the pose of a wrong target.}
    \label{fig:failure}
\end{figure}

\paragraph{Visualization of Shared Feature Space.}
As shown in Fig~\ref{fig:feature}, embeddings of different modalities are aligned with each other and uniformly distributed in the shared feature space. The visualization is based on t-SNE\cite{van2008tsne}.
\paragraph{Failure cases.} As shown in Figure~\ref{fig:failure}, the image branch of UniHPE fails in the case of large occlusion or low-quality RGB input scenarios. UniHPE is trained on Human3.6M and MPI-INF-3DHP with only one target per frame and a limited amount of occlusion.


\section{Limitation and Future Work}
Moreover, our proposed UniHPE relies on the availability of large-scale and diverse datasets for pre-training using contrastive learning. While the approach aligns RGB image, 2D, and 3D human pose representations in a shared feature space, its performance is likely contingent on the diversity and amount of data it is trained on. This could bring challenges in environments where such rich datasets are not available, potentially limiting the model's ability to generalize to new or rare poses. For our future work, adopting SMPL model may help the pipeline better utilize semi-supervised learning by taking advantage of 2D human pose estimation datasets. Meanwhile, the aligned embeddings and pre-trained model from UniHPE may benefit more downstream tasks, like human re-identification, action recognition, \etc. 

\vspace{-0.2cm}
\section{Conclusion}
In conclusion, the UniHPE framework represents a significant step forward in unified human pose estimation by integrating the strengths of lifting and image-based methods within a single model. It leverages contrastive learning to align embeddings across modalities in a shared feature space, enhancing the model's accuracy in pose estimation tasks. Despite its potential limitations in data requirements and computational intensity, UniHPE sets a promising direction for future research, particularly in improving generalization capabilities and a unified end-to-end pipeline. The framework's achievements on benchmark datasets like Human3.6M and 3DPW underscore its potential, paving the way for advancements in applications across multiple domains such as motion analysis, augmented reality, and human-computer interaction.

\newpage
{
    \small
    \bibliographystyle{ieeenat_fullname}
    \bibliography{main}

\begin{thebibliography}{66}
\providecommand{\natexlab}[1]{#1}
\providecommand{\url}[1]{\texttt{#1}}
\expandafter\ifx\csname urlstyle\endcsname\relax
  \providecommand{\doi}[1]{doi: #1}\else
  \providecommand{\doi}{doi: \begingroup \urlstyle{rm}\Url}\fi

\bibitem[Andriluka et~al.(2018)Andriluka, Iqbal, Insafutdinov, Pishchulin, Milan, Gall, and Schiele]{andriluka2018posetrack}
Mykhaylo Andriluka, Umar Iqbal, Eldar Insafutdinov, Leonid Pishchulin, Anton Milan, Juergen Gall, and Bernt Schiele.
\newblock Posetrack: A benchmark for human pose estimation and tracking.
\newblock In \emph{Proceedings of the IEEE conference on computer vision and pattern recognition}, pages 5167--5176, 2018.

\bibitem[Arnab et~al.(2019)Arnab, Doersch, and Zisserman]{arnab2019exploiting}
Anurag Arnab, Carl Doersch, and Andrew Zisserman.
\newblock Exploiting temporal context for 3d human pose estimation in the wild.
\newblock In \emph{Proceedings of the IEEE/CVF Conference on Computer Vision and Pattern Recognition}, pages 3395--3404, 2019.

\bibitem[Bogo et~al.(2016)Bogo, Kanazawa, Lassner, Gehler, Romero, and Black]{bogo2016smplify}
Federica Bogo, Angjoo Kanazawa, Christoph Lassner, Peter Gehler, Javier Romero, and Michael~J Black.
\newblock Keep it smpl: Automatic estimation of 3d human pose and shape from a single image.
\newblock In \emph{Computer Vision--ECCV 2016: 14th European Conference, Amsterdam, The Netherlands, October 11-14, 2016, Proceedings, Part V 14}, pages 561--578. Springer, 2016.

\bibitem[Chai et~al.(2023)Chai, Jiang, Hwang, and Wang]{chai2023global}
Wenhao Chai, Zhongyu Jiang, Jenq-Neng Hwang, and Gaoang Wang.
\newblock Global adaptation meets local generalization: Unsupervised domain adaptation for 3d human pose estimation.
\newblock \emph{arXiv preprint arXiv:2303.16456}, 2023.

\bibitem[Ci et~al.(2023)Ci, Wu, Zhu, Ma, Dong, Zhong, and Wang]{ci2023gfpose}
Hai Ci, Mingdong Wu, Wentao Zhu, Xiaoxuan Ma, Hao Dong, Fangwei Zhong, and Yizhou Wang.
\newblock Gfpose: Learning 3d human pose prior with gradient fields.
\newblock In \emph{Proceedings of the IEEE/CVF Conference on Computer Vision and Pattern Recognition}, pages 4800--4810, 2023.

\bibitem[Doersch and Zisserman(2019)]{doersch2019sim2real}
Carl Doersch and Andrew Zisserman.
\newblock Sim2real transfer learning for 3d human pose estimation: motion to the rescue.
\newblock \emph{Advances in Neural Information Processing Systems}, 32, 2019.

\bibitem[Duan et~al.(2022)Duan, Wang, Chen, and Lin]{duan2022pyskl}
Haodong Duan, Jiaqi Wang, Kai Chen, and Dahua Lin.
\newblock Pyskl: Towards good practices for skeleton action recognition.
\newblock In \emph{Proceedings of the 30th ACM International Conference on Multimedia}, pages 7351--7354, 2022.

\bibitem[Georgakis et~al.(2020{\natexlab{a}})Georgakis, Li, Karanam, Chen, Ko{\v{s}}eck{\'a}, and Wu]{georgakis2020hierarchical}
Georgios Georgakis, Ren Li, Srikrishna Karanam, Terrence Chen, Jana Ko{\v{s}}eck{\'a}, and Ziyan Wu.
\newblock Hierarchical kinematic human mesh recovery.
\newblock In \emph{Computer Vision--ECCV 2020: 16th European Conference, Glasgow, UK, August 23--28, 2020, Proceedings, Part XVII 16}, pages 768--784. Springer, 2020{\natexlab{a}}.

\bibitem[Georgakis et~al.(2020{\natexlab{b}})Georgakis, Li, Karanam, Chen, Ko{\v{s}}eck{\'a}, and Wu]{georgakis2020hkmr}
Georgios Georgakis, Ren Li, Srikrishna Karanam, Terrence Chen, Jana Ko{\v{s}}eck{\'a}, and Ziyan Wu.
\newblock Hierarchical kinematic human mesh recovery.
\newblock In \emph{Computer Vision--ECCV 2020: 16th European Conference, Glasgow, UK, August 23--28, 2020, Proceedings, Part XVII 16}, pages 768--784. Springer, 2020{\natexlab{b}}.

\bibitem[Gholami et~al.(2022)Gholami, Wandt, Rhodin, Ward, and Wang]{gholami2022adaptpose}
Mohsen Gholami, Bastian Wandt, Helge Rhodin, Rabab Ward, and Z~Jane Wang.
\newblock Adaptpose: Cross-dataset adaptation for 3d human pose estimation by learnable motion generation.
\newblock In \emph{Proceedings of the IEEE/CVF Conference on Computer Vision and Pattern Recognition}, pages 13075--13085, 2022.

\bibitem[Girdhar et~al.(2023)Girdhar, El-Nouby, Liu, Singh, Alwala, Joulin, and Misra]{girdhar2023imagebind}
Rohit Girdhar, Alaaeldin El-Nouby, Zhuang Liu, Mannat Singh, Kalyan~Vasudev Alwala, Armand Joulin, and Ishan Misra.
\newblock Imagebind: One embedding space to bind them all.
\newblock In \emph{Proceedings of the IEEE/CVF Conference on Computer Vision and Pattern Recognition}, pages 15180--15190, 2023.

\bibitem[Gong et~al.(2021)Gong, Zhang, and Feng]{gong2021poseaug}
Kehong Gong, Jianfeng Zhang, and Jiashi Feng.
\newblock Poseaug: A differentiable pose augmentation framework for 3d human pose estimation.
\newblock In \emph{Proceedings of the IEEE/CVF conference on computer vision and pattern recognition}, pages 8575--8584, 2021.

\bibitem[Guler and Kokkinos(2019)]{guler2019holopose}
Riza~Alp Guler and Iasonas Kokkinos.
\newblock Holopose: Holistic 3d human reconstruction in-the-wild.
\newblock In \emph{Proceedings of the IEEE/CVF Conference on Computer Vision and Pattern Recognition}, pages 10884--10894, 2019.

\bibitem[Guzhov et~al.(2022)Guzhov, Raue, Hees, and Dengel]{guzhov2022audioclip}
Andrey Guzhov, Federico Raue, J{\"o}rn Hees, and Andreas Dengel.
\newblock Audioclip: Extending clip to image, text and audio.
\newblock In \emph{ICASSP 2022-2022 IEEE International Conference on Acoustics, Speech and Signal Processing (ICASSP)}, pages 976--980. IEEE, 2022.

\bibitem[Ho et~al.(2020)Ho, Jain, and Abbeel]{ddpm}
Jonathan Ho, Ajay Jain, and Pieter Abbeel.
\newblock Denoising diffusion probabilistic models.
\newblock \emph{Advances in neural information processing systems}, 33:\penalty0 6840--6851, 2020.

\bibitem[Hu et~al.(2018)Hu, Jiang, Ding, Mu, and Hall]{hu2018vgpn}
Jun Hu, Zhongyu Jiang, Xionghao Ding, Taijiang Mu, and Peter Hall.
\newblock Vgpn: Voice-guided pointing robot navigation for humans.
\newblock In \emph{2018 IEEE International Conference on Robotics and Biomimetics (ROBIO)}, pages 1107--1112. IEEE, 2018.

\bibitem[Ionescu et~al.(2013)Ionescu, Papava, Olaru, and Sminchisescu]{ionescu2013h36m}
Catalin Ionescu, Dragos Papava, Vlad Olaru, and Cristian Sminchisescu.
\newblock Human3. 6m: Large scale datasets and predictive methods for 3d human sensing in natural environments.
\newblock \emph{IEEE transactions on pattern analysis and machine intelligence}, 36\penalty0 (7):\penalty0 1325--1339, 2013.

\bibitem[Jiang et~al.(2022)Jiang, Ji, Menaker, and Hwang]{jiang2022golfpose}
Zhongyu Jiang, Haorui Ji, Samuel Menaker, and Jenq-Neng Hwang.
\newblock Golfpose: Golf swing analyses with a monocular camera based human pose estimation.
\newblock In \emph{2022 IEEE International Conference on Multimedia and Expo Workshops (ICMEW)}, pages 1--6. IEEE, 2022.

\bibitem[Jiang et~al.(2023)Jiang, Zhou, Li, Chai, Yang, and Hwang]{jiang2023zedo}
Zhongyu Jiang, Zhuoran Zhou, Lei Li, Wenhao Chai, Cheng-Yen Yang, and Jenq-Neng Hwang.
\newblock Back to optimization: Diffusion-based zero-shot 3d human pose estimation.
\newblock \emph{arXiv preprint arXiv:2307.03833}, 2023.

\bibitem[Kanazawa et~al.(2018)Kanazawa, Black, Jacobs, and Malik]{kanazawa2018hmr}
Angjoo Kanazawa, Michael~J Black, David~W Jacobs, and Jitendra Malik.
\newblock End-to-end recovery of human shape and pose.
\newblock In \emph{Proceedings of the IEEE conference on computer vision and pattern recognition}, pages 7122--7131, 2018.

\bibitem[Kanazawa et~al.(2019)Kanazawa, Zhang, Felsen, and Malik]{kanazawa2019learning}
Angjoo Kanazawa, Jason~Y Zhang, Panna Felsen, and Jitendra Malik.
\newblock Learning 3d human dynamics from video.
\newblock In \emph{Proceedings of the IEEE/CVF conference on computer vision and pattern recognition}, pages 5614--5623, 2019.

\bibitem[Kocabas et~al.(2020)Kocabas, Athanasiou, and Black]{kocabas2020vibe}
Muhammed Kocabas, Nikos Athanasiou, and Michael~J Black.
\newblock Vibe: Video inference for human body pose and shape estimation.
\newblock In \emph{Proceedings of the IEEE/CVF conference on computer vision and pattern recognition}, pages 5253--5263, 2020.

\bibitem[Kocabas et~al.(2021)Kocabas, Huang, Hilliges, and Black]{kocabas2021pare}
Muhammed Kocabas, Chun-Hao~P Huang, Otmar Hilliges, and Michael~J Black.
\newblock Pare: Part attention regressor for 3d human body estimation.
\newblock In \emph{Proceedings of the IEEE/CVF International Conference on Computer Vision}, pages 11127--11137, 2021.

\bibitem[Kolotouros et~al.(2019{\natexlab{a}})Kolotouros, Pavlakos, Black, and Daniilidis]{kolotouros2019spin}
Nikos Kolotouros, Georgios Pavlakos, Michael~J Black, and Kostas Daniilidis.
\newblock Learning to reconstruct 3d human pose and shape via model-fitting in the loop.
\newblock In \emph{ICCV}, 2019{\natexlab{a}}.

\bibitem[Kolotouros et~al.(2019{\natexlab{b}})Kolotouros, Pavlakos, and Daniilidis]{kolotouros2019cmr}
Nikos Kolotouros, Georgios Pavlakos, and Kostas Daniilidis.
\newblock Convolutional mesh regression for single-image human shape reconstruction.
\newblock In \emph{CVPR}, 2019{\natexlab{b}}.

\bibitem[Li et~al.(2023)Li, Bian, Xu, Chen, Yang, and Lu]{li2023hybrik}
Jiefeng Li, Siyuan Bian, Chao Xu, Zhicun Chen, Lixin Yang, and Cewu Lu.
\newblock Hybrik-x: Hybrid analytical-neural inverse kinematics for whole-body mesh recovery.
\newblock \emph{arXiv preprint arXiv:2304.05690}, 2023.

\bibitem[Li et~al.(2022{\natexlab{a}})Li, Liu, Ding, Liu, Wang, and Yang]{li2022exploiting}
Wenhao Li, Hong Liu, Runwei Ding, Mengyuan Liu, Pichao Wang, and Wenming Yang.
\newblock Exploiting temporal contexts with strided transformer for 3d human pose estimation.
\newblock \emph{IEEE Transactions on Multimedia}, 25:\penalty0 1282--1293, 2022{\natexlab{a}}.

\bibitem[Li et~al.(2022{\natexlab{b}})Li, Liu, Zhang, Xu, and Yan]{li2022cliff}
Zhihao Li, Jianzhuang Liu, Zhensong Zhang, Songcen Xu, and Youliang Yan.
\newblock Cliff: Carrying location information in full frames into human pose and shape estimation.
\newblock In \emph{European Conference on Computer Vision}, pages 590--606. Springer, 2022{\natexlab{b}}.

\bibitem[Lin et~al.(2022)Lin, Geng, Zhang, Gao, de~Melo, Wang, Dai, Qiao, and Li]{lin2022frozen}
Ziyi Lin, Shijie Geng, Renrui Zhang, Peng Gao, Gerard de Melo, Xiaogang Wang, Jifeng Dai, Yu Qiao, and Hongsheng Li.
\newblock Frozen clip models are efficient video learners.
\newblock In \emph{European Conference on Computer Vision}, pages 388--404. Springer, 2022.

\bibitem[Loper et~al.(2015)Loper, Mahmood, Romero, Pons-Moll, and Black]{SMPL:2015}
Matthew Loper, Naureen Mahmood, Javier Romero, Gerard Pons-Moll, and Michael~J. Black.
\newblock {SMPL}: A skinned multi-person linear model.
\newblock \emph{ACM Trans. Graphics (Proc. SIGGRAPH Asia)}, 34\penalty0 (6):\penalty0 248:1--248:16, 2015.

\bibitem[Luo et~al.(2022)Luo, Ji, Zhong, Chen, Lei, Duan, and Li]{luo2022clip4clip}
Huaishao Luo, Lei Ji, Ming Zhong, Yang Chen, Wen Lei, Nan Duan, and Tianrui Li.
\newblock Clip4clip: An empirical study of clip for end to end video clip retrieval and captioning.
\newblock pages 293--304. Elsevier, 2022.

\bibitem[Luvizon et~al.(2018)Luvizon, Picard, and Tabia]{luvizon20182d}
Diogo~C Luvizon, David Picard, and Hedi Tabia.
\newblock 2d/3d pose estimation and action recognition using multitask deep learning.
\newblock In \emph{Proceedings of the IEEE conference on computer vision and pattern recognition}, pages 5137--5146, 2018.

\bibitem[Martinez et~al.(2017)Martinez, Hossain, Romero, and Little]{martinez2017simplebaseline}
Julieta Martinez, Rayat Hossain, Javier Romero, and James~J Little.
\newblock A simple yet effective baseline for 3d human pose estimation.
\newblock In \emph{Proceedings of the IEEE international conference on computer vision}, pages 2640--2649, 2017.

\bibitem[Mehta et~al.(2017)Mehta, Rhodin, Casas, Fua, Sotnychenko, Xu, and Theobalt]{mono-3dhp2017}
Dushyant Mehta, Helge Rhodin, Dan Casas, Pascal Fua, Oleksandr Sotnychenko, Weipeng Xu, and Christian Theobalt.
\newblock Monocular 3d human pose estimation in the wild using improved cnn supervision.
\newblock In \emph{3D Vision (3DV), 2017 Fifth International Conference on}. IEEE, 2017.

\bibitem[Omran et~al.(2018)Omran, Lassner, Pons-Moll, Gehler, and Schiele]{omran2018nbf}
Mohamed Omran, Christoph Lassner, Gerard Pons-Moll, Peter~V. Gehler, and Bernt Schiele.
\newblock Neural body fitting: Unifying deep learning and model-based human pose and shape estimation.
\newblock 2018.

\bibitem[Paszke et~al.(2019)Paszke, Gross, Massa, Lerer, Bradbury, Chanan, Killeen, Lin, Gimelshein, Antiga, et~al.]{paszke2019pytorch}
Adam Paszke, Sam Gross, Francisco Massa, Adam Lerer, James Bradbury, Gregory Chanan, Trevor Killeen, Zeming Lin, Natalia Gimelshein, Luca Antiga, et~al.
\newblock Pytorch: An imperative style, high-performance deep learning library.
\newblock \emph{Advances in neural information processing systems}, 32, 2019.

\bibitem[Pavlakos et~al.(2017)Pavlakos, Zhou, Derpanis, and Daniilidis]{pavlakos2017coarse}
Georgios Pavlakos, Xiaowei Zhou, Konstantinos~G Derpanis, and Kostas Daniilidis.
\newblock Coarse-to-fine volumetric prediction for single-image 3d human pose.
\newblock In \emph{Proceedings of the IEEE conference on computer vision and pattern recognition}, pages 7025--7034, 2017.

\bibitem[Pavlakos et~al.(2018)Pavlakos, Zhu, Zhou, and Daniilidis]{pavlakos2018learning}
Georgios Pavlakos, Luyang Zhu, Xiaowei Zhou, and Kostas Daniilidis.
\newblock Learning to estimate 3d human pose and shape from a single color image.
\newblock In \emph{Proceedings of the IEEE conference on computer vision and pattern recognition}, pages 459--468, 2018.

\bibitem[Pavllo et~al.(2019)Pavllo, Feichtenhofer, Grangier, and Auli]{pavllo2019videopose3d}
Dario Pavllo, Christoph Feichtenhofer, David Grangier, and Michael Auli.
\newblock 3d human pose estimation in video with temporal convolutions and semi-supervised training.
\newblock In \emph{Proceedings of the IEEE/CVF conference on computer vision and pattern recognition}, pages 7753--7762, 2019.

\bibitem[Radford et~al.(2021)Radford, Kim, Hallacy, Ramesh, Goh, Agarwal, Sastry, Askell, Mishkin, Clark, et~al.]{radford2021clip}
Alec Radford, Jong~Wook Kim, Chris Hallacy, Aditya Ramesh, Gabriel Goh, Sandhini Agarwal, Girish Sastry, Amanda Askell, Pamela Mishkin, Jack Clark, et~al.
\newblock Learning transferable visual models from natural language supervision.
\newblock In \emph{International Conference on Machine Learning}, pages 8748--8763. PMLR, 2021.

\bibitem[Shahroudy et~al.(2016)Shahroudy, Liu, Ng, and Wang]{shahroudy2016nturgbd}
Amir Shahroudy, Jun Liu, Tian-Tsong Ng, and Gang Wang.
\newblock Ntu rgb+d: A large scale dataset for 3d human activity analysis.
\newblock In \emph{Proceedings of the IEEE conference on computer vision and pattern recognition}, pages 1010--1019, 2016.

\bibitem[Snower et~al.(2020)Snower, Kadav, Lai, and Graf]{snower2020keytrack}
Michael Snower, Asim Kadav, Farley Lai, and Hans~Peter Graf.
\newblock 15 keypoints is all you need.
\newblock In \emph{Proceedings of the IEEE/CVF Conference on Computer Vision and Pattern Recognition}, pages 6738--6748, 2020.

\bibitem[Song et~al.(2020{\natexlab{a}})Song, Meng, and Ermon]{ddim}
Jiaming Song, Chenlin Meng, and Stefano Ermon.
\newblock Denoising diffusion implicit models.
\newblock \emph{arXiv preprint arXiv:2010.02502}, 2020{\natexlab{a}}.

\bibitem[Song and Ermon(2019)]{SMN}
Yang Song and Stefano Ermon.
\newblock Generative modeling by estimating gradients of the data distribution.
\newblock In \emph{Advances in Neural Information Processing Systems}. Curran Associates, Inc., 2019.

\bibitem[Song et~al.(2020{\natexlab{b}})Song, Sohl-Dickstein, Kingma, Kumar, Ermon, and Poole]{song2020score}
Yang Song, Jascha Sohl-Dickstein, Diederik~P Kingma, Abhishek Kumar, Stefano Ermon, and Ben Poole.
\newblock Score-based generative modeling through stochastic differential equations.
\newblock \emph{arXiv preprint arXiv:2011.13456}, 2020{\natexlab{b}}.

\bibitem[Sun et~al.(2019)Sun, Ye, Liu, Gao, Fu, and Mei]{sun2019dsd}
Yu Sun, Yun Ye, Wu Liu, Wenpeng Gao, Yili Fu, and Tao Mei.
\newblock Human mesh recovery from monocular images via a skeleton-disentangled representation.
\newblock In \emph{Proceedings of the IEEE/CVF international conference on computer vision}, pages 5349--5358, 2019.

\bibitem[Sun et~al.(2021)Sun, Bao, Liu, Fu, Michael~J., and Mei]{sun2021ROMP}
Yu Sun, Qian Bao, Wu Liu, Yili Fu, Black Michael~J., and Tao Mei.
\newblock {Monocular, One-stage, Regression of Multiple 3D People}.
\newblock In \emph{ICCV}, 2021.

\bibitem[Sun et~al.(2022)Sun, Liu, Bao, Fu, Mei, and Black]{sun2022BEV}
Yu Sun, Wu Liu, Qian Bao, Yili Fu, Tao Mei, and Michael~J Black.
\newblock {Putting People in their Place: Monocular Regression of 3D People in Depth}.
\newblock In \emph{CVPR}, 2022.

\bibitem[Van~der Maaten and Hinton(2008)]{van2008tsne}
Laurens Van~der Maaten and Geoffrey Hinton.
\newblock Visualizing data using t-sne.
\newblock \emph{Journal of machine learning research}, 9\penalty0 (11), 2008.

\bibitem[Vaswani et~al.(2017)Vaswani, Shazeer, Parmar, Uszkoreit, Jones, Gomez, Kaiser, and Polosukhin]{vaswani2017attention}
Ashish Vaswani, Noam Shazeer, Niki Parmar, Jakob Uszkoreit, Llion Jones, Aidan~N Gomez, {\L}ukasz Kaiser, and Illia Polosukhin.
\newblock Attention is all you need.
\newblock \emph{Advances in neural information processing systems}, 30, 2017.

\bibitem[Von~Marcard et~al.(2018)Von~Marcard, Henschel, Black, Rosenhahn, and Pons-Moll]{von20183dpw}
Timo Von~Marcard, Roberto Henschel, Michael~J Black, Bodo Rosenhahn, and Gerard Pons-Moll.
\newblock Recovering accurate 3d human pose in the wild using imus and a moving camera.
\newblock In \emph{Proceedings of the European conference on computer vision (ECCV)}, pages 601--617, 2018.

\bibitem[Wang et~al.(2020)Wang, Sun, Cheng, Jiang, Deng, Zhao, Liu, Mu, Tan, Wang, et~al.]{wang2020deep}
Jingdong Wang, Ke Sun, Tianheng Cheng, Borui Jiang, Chaorui Deng, Yang Zhao, Dong Liu, Yadong Mu, Mingkui Tan, Xinggang Wang, et~al.
\newblock Deep high-resolution representation learning for visual recognition.
\newblock \emph{IEEE transactions on pattern analysis and machine intelligence}, 43\penalty0 (10):\penalty0 3349--3364, 2020.

\bibitem[Wang et~al.(2021)Wang, Tan, Zhen, Xu, Zheng, He, and Shao]{wang2021deep}
Jinbao Wang, Shujie Tan, Xiantong Zhen, Shuo Xu, Feng Zheng, Zhenyu He, and Ling Shao.
\newblock Deep 3d human pose estimation: A review.
\newblock \emph{Computer Vision and Image Understanding}, 210:\penalty0 103225, 2021.

\bibitem[Wang et~al.(2018)Wang, Zhao, and Ji]{wang2018hci}
Kang Wang, Rui Zhao, and Qiang Ji.
\newblock Human computer interaction with head pose, eye gaze and body gestures.
\newblock In \emph{2018 13th IEEE International Conference on Automatic Face \& Gesture Recognition (FG 2018)}, pages 789--789. IEEE, 2018.

\bibitem[Weng et~al.(2022)Weng, Curless, Srinivasan, Barron, and Kemelmacher-Shlizerman]{weng2022humannerf}
Chung-Yi Weng, Brian Curless, Pratul~P Srinivasan, Jonathan~T Barron, and Ira Kemelmacher-Shlizerman.
\newblock Humannerf: Free-viewpoint rendering of moving people from monocular video.
\newblock In \emph{Proceedings of the IEEE/CVF conference on computer vision and pattern Recognition}, pages 16210--16220, 2022.

\bibitem[Xu et~al.(2019)Xu, Zhu, and Tung]{xu2019denserac}
Yuanlu Xu, Song-Chun Zhu, and Tony Tung.
\newblock Denserac: Joint 3d pose and shape estimation by dense render-and-compare.
\newblock In \emph{Proceedings of the IEEE/CVF International Conference on Computer Vision}, pages 7760--7770, 2019.

\bibitem[Xu et~al.(2022)Xu, Zhang, Zhang, and Tao]{xu2022vitpose}
Yufei Xu, Jing Zhang, Qiming Zhang, and Dacheng Tao.
\newblock Vitpose: Simple vision transformer baselines for human pose estimation.
\newblock \emph{Advances in Neural Information Processing Systems}, 35:\penalty0 38571--38584, 2022.

\bibitem[Zeng et~al.(2020)Zeng, Ouyang, Luo, Liu, and Wang]{zeng2020decomr}
Wang Zeng, Wanli Ouyang, Ping Luo, Wentao Liu, and Xiaogang Wang.
\newblock 3d human mesh regression with dense correspondence.
\newblock In \emph{Proceedings of the IEEE/CVF conference on computer vision and pattern recognition}, pages 7054--7063, 2020.

\bibitem[Zhang et~al.(2021)Zhang, Tian, Zhou, Ouyang, Liu, Wang, and Sun]{zhang2021pymaf}
Hongwen Zhang, Yating Tian, Xinchi Zhou, Wanli Ouyang, Yebin Liu, Limin Wang, and Zhenan Sun.
\newblock Pymaf: 3d human pose and shape regression with pyramidal mesh alignment feedback loop.
\newblock In \emph{Proceedings of the IEEE/CVF International Conference on Computer Vision}, pages 11446--11456, 2021.

\bibitem[Zhang et~al.(2023{\natexlab{a}})Zhang, Tian, Zhang, Li, An, Sun, and Liu]{zhang2023pymafx}
Hongwen Zhang, Yating Tian, Yuxiang Zhang, Mengcheng Li, Liang An, Zhenan Sun, and Yebin Liu.
\newblock Pymaf-x: Towards well-aligned full-body model regression from monocular images.
\newblock \emph{IEEE Transactions on Pattern Analysis and Machine Intelligence}, 2023{\natexlab{a}}.

\bibitem[Zhang et~al.(2022)Zhang, Tu, Yang, Chen, and Yuan]{zhang2022mixste}
Jinlu Zhang, Zhigang Tu, Jianyu Yang, Yujin Chen, and Junsong Yuan.
\newblock Mixste: Seq2seq mixed spatio-temporal encoder for 3d human pose estimation in video.
\newblock In \emph{Proceedings of the IEEE/CVF conference on computer vision and pattern recognition}, pages 13232--13242, 2022.

\bibitem[Zhang et~al.(2023{\natexlab{b}})Zhang, Chai, Jiang, Ye, Song, Hwang, and Wang]{zhang2023mpm}
Zhenyu Zhang, Wenhao Chai, Zhongyu Jiang, Tian Ye, Mingli Song, Jenq-Neng Hwang, and Gaoang Wang.
\newblock Mpm: A unified 2d-3d human pose representation via masked pose modeling.
\newblock \emph{arXiv preprint arXiv:2306.17201}, 2023{\natexlab{b}}.

\bibitem[Zhao et~al.(2019)Zhao, Peng, Tian, Kapadia, and Metaxas]{zhao2019semgcn}
Long Zhao, Xi Peng, Yu Tian, Mubbasir Kapadia, and Dimitris~N Metaxas.
\newblock Semantic graph convolutional networks for 3d human pose regression.
\newblock In \emph{Proceedings of the IEEE/CVF conference on computer vision and pattern recognition}, pages 3425--3435, 2019.

\bibitem[Zhao et~al.(2023)Zhao, Chai, Hao, Hu, Wang, Cao, Song, Hwang, and Wang]{zhao2023survey}
Zhonghan Zhao, Wenhao Chai, Shengyu Hao, Wenhao Hu, Guanhong Wang, Shidong Cao, Mingli Song, Jenq-Neng Hwang, and Gaoang Wang.
\newblock A survey of deep learning in sports applications: Perception, comprehension, and decision.
\newblock \emph{arXiv preprint arXiv:2307.03353}, 2023.

\bibitem[Zheng et~al.(2021)Zheng, Zhu, Mendieta, Yang, Chen, and Ding]{zheng2021poseformer}
Ce Zheng, Sijie Zhu, Matias Mendieta, Taojiannan Yang, Chen Chen, and Zhengming Ding.
\newblock 3d human pose estimation with spatial and temporal transformers.
\newblock In \emph{Proceedings of the IEEE/CVF International Conference on Computer Vision}, pages 11656--11665, 2021.

\bibitem[Zhou et~al.(2023)Zhou, Jiang, Chai, Yang, Li, and Hwang]{zhou2023zedoi}
Zhuoran Zhou, Zhongyu Jiang, Wenhao Chai, Cheng-Yen Yang, Lei Li, and Jenq-Neng Hwang.
\newblock Efficient domain adaptation via generative prior for 3d infant pose estimation, 2023.

\end{thebibliography}
}

\clearpage
\setcounter{page}{1}
\maketitlesupplementary
\setcounter{section}{0}
\renewcommand{\thesection}{\Alph{section}}

\begin{figure*}[b]
    \centering
    \includegraphics[width=\linewidth]{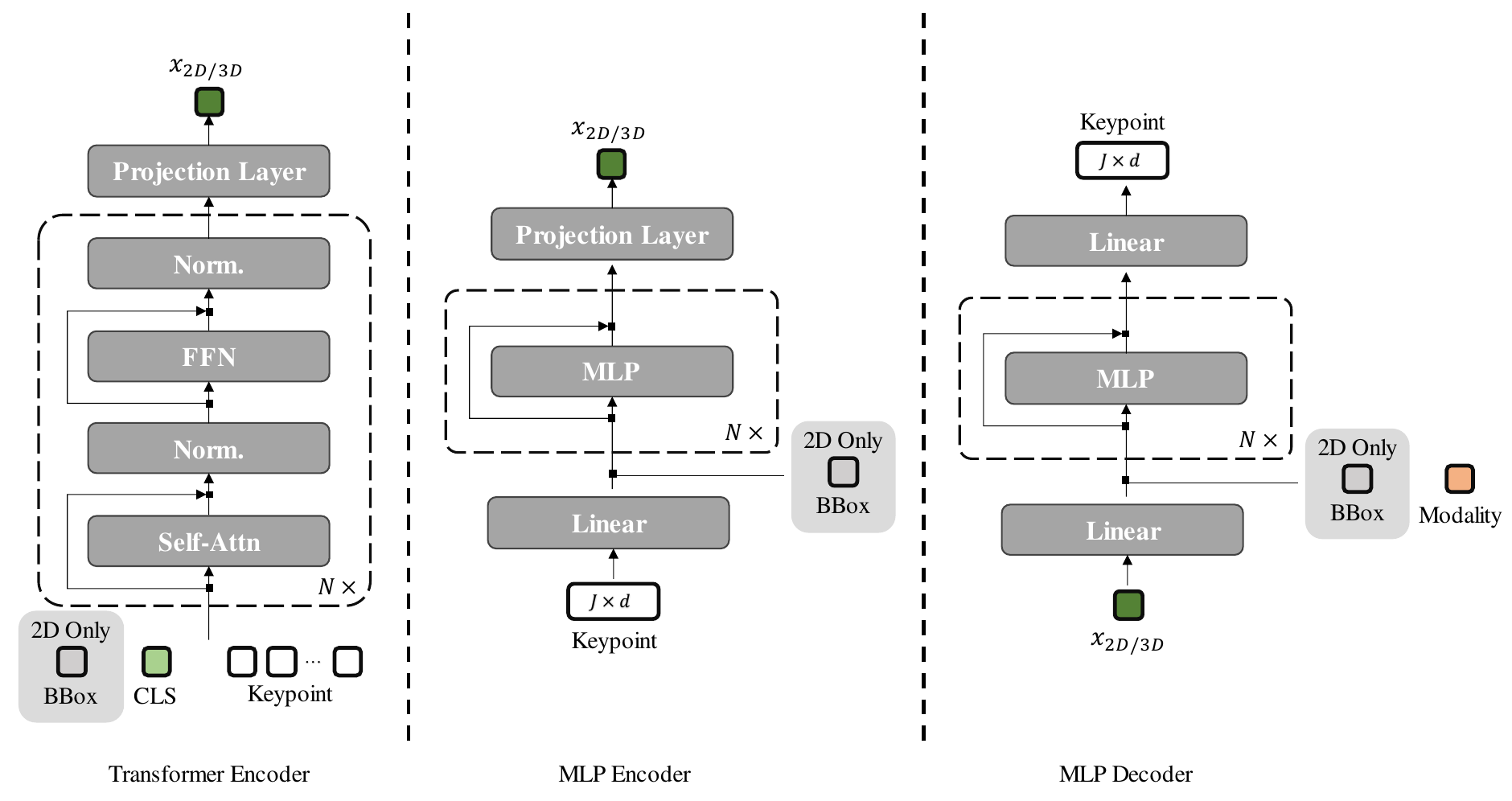}
    \caption{The architectures of the transformer based encoder and MLP based encoder and decoder. }
    \label{fig:supp_arch}
\end{figure*}

\section{Details of Singular Value based Loss}
\label{supp:singular}

In this paper, we propose a singular value based contrastive learning loss to align more than two modalities simultaneously. In this section, more details of the singular value based loss are discussed, including training speed and hyperparameters.

\begin{table}[b]
\centering
\newcommand{\R}[1]{~\scriptsize{\textcolor{red}{(#1)}}}
\resizebox{0.9\linewidth}{!}{
\begin{tabular}{l|c}
\toprule
Loss & Per Batch Time (s) \\
\midrule
$\mathcal{L}_{pair}$ & 1.77 \\
$\mathcal{L}_{pair} + \mathcal{L}_{triplet}$ (singular value) & 2.64\R{+49.1\%} \\
\midrule
\rowcolor[gray]{0.9}
$\mathcal{L}_{pair} + \mathcal{L}_{triplet}$ (eigenvalue) & 2.01\R{+13.5\%} \\
\bottomrule
\end{tabular}
}
\caption{Per batch training speed with Nvidia A100. The proposed $\mathcal{L}_{triplet}$ does not significantly increase the training complexity and can be easily adopted to other multi-modal alignment training pipelines}
\label{tab:training-speed}
\end{table}

\vspace{-0.2cm}

\paragraph{Training Speed.} We compare the training speed difference per batch on the Nvidia A100 with three different loss setup: $\mathcal{L}_{pair}$ only, $\mathcal{L}_{pair} + \mathcal{L}_{triplet}$ with singular values, and $\mathcal{L}_{pair} + \mathcal{L}_{triplet}$ with eigenvalues. As shown in Table~\ref{tab:training-speed}, the eigenvalue based $\mathcal{L}_{triplet}$ does not significantly increase the training complexity and can be easily adopted to other multi-modal alignment training pipelines.

\vspace{-0.2cm}

\paragraph{Hyperparameters.} In each batch $B$, for every triplet of data, we randomly sample $B - 1$ negative triplets. Therefore, eigenvalues of $B \times B$ matrices are calculated. In our experiment, the weight, $\alpha$, of $\mathcal{L}_{triplet}$ is $1$ and the learnable temperature, $\tau$, is shared with $\mathcal{L}_{pair}$.

\section{Architecture Details}
\label{supp:arch}

In this section, we introduce the detailed architectures of 2D and 3D pose encoders and decoders.

\paragraph{The bounding box token.} According to \cite{li2022cliff}, without bounding box information, different 3D human poses may have exact the same 2D human poses in the cropped bounding box. Therefore, in our 2D encoder and 2D decoder, there is an additional bounding box token to provide bounding box information.

\subsection{MLP Encoders and Decoders}

We implement MLP with residual blocks as our MLP encoders and decoders, as shown in Fig~\ref{fig:supp_arch}. The MLP encoder and decoder are constructed with $3$ MLP with residual blocks. As mentioned in Sec 3, there is a modality token, $\mathcal{M}$, which indicates the input modality to further boost the performance. 

\subsection{Transformer Encoders}
For pose encoding, our 2D and 3D transformer encoders share similar architectures, except the 2D encoder takes an additional bounding box token as input to solve the projection ambiguity. A $\left[CLS\right]$ token, $J$ keypoint tokens projected from 2D and 3D keypoints, and an additional bounding box token for the 2D encoder are the input of the encoder. After three transformer encoder layers with multi-head attention, the embedding of the $\left[CLS\right]$ token is projected to the shared feature space with dimension as $1024$.

\subsection{Diffusion Decoders}

We apply the same architecture as GFPose\cite{ci2023gfpose} and Score Matching Network~\cite{song2020score} as our diffusion decoders with an additional bounding box token and modality token. The decoder supposedly recovers accurate keypoints from random Gaussian noise by giving the previous resulting tokens as conditions. We set the timestep, $t$, in $\left(0, 1\right]$ and uniformly sample the timestep $1000$ times in training and inference. Different from the original GFPose, our keypoints are defined as pelvis-relative in the camera coordinate. 

\begin{figure*}[t]
    \centering
    \includegraphics[width=\linewidth]{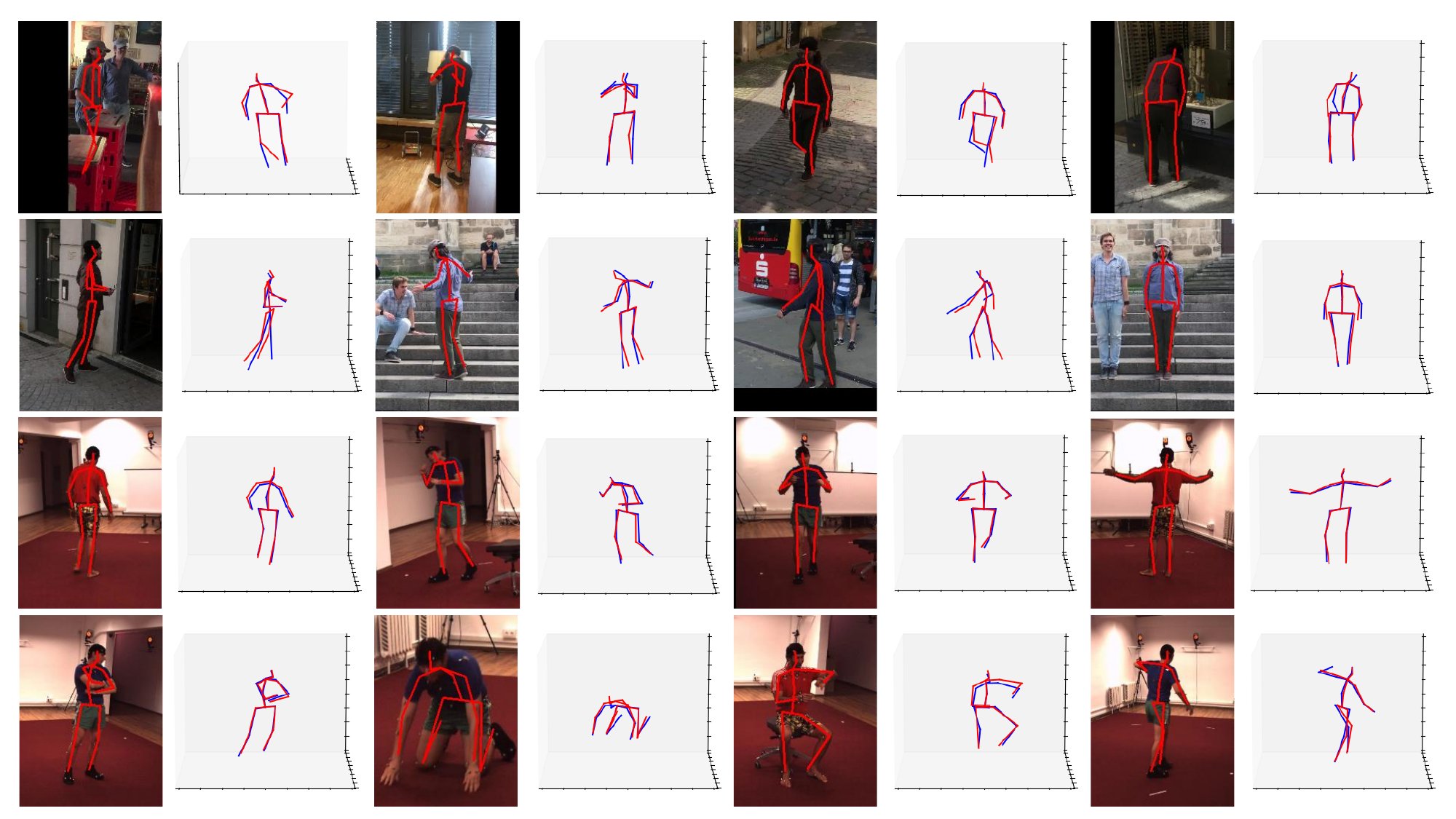}
    \caption{More visualization of our method. The first two rows are from 3DPW dataset, while the last two rows are from Human3.6M dataset. The red skeleton is the ground truth, and the blue one is estimated from the image branch.}
    \label{fig:supp_arch}
\end{figure*}


\end{document}